\documentclass[journal]{IEEEtran}
\usepackage{times}
\usepackage{epsfig}
\usepackage{graphicx}
\usepackage{amsmath}
\usepackage{amssymb,pifont}

\usepackage{multirow}
\usepackage{booktabs}
\usepackage{bm}
\usepackage{colortbl}

\usepackage{cite}
\usepackage{makecell}
\usepackage{boldline}
\setcellgapes{3pt}
\usepackage{arydshln}
\usepackage{tabu}
\usepackage{subfigure}
\usepackage{balance}
\usepackage{xcolor}
\usepackage[breaklinks=true,bookmarks=false]{hyperref}

\usepackage{cleveref}
\crefformat{section}{\S#2#1#3} 
\crefformat{subsection}{\S#2#1#3}
\crefformat{subsubsection}{\S#2#1#3}

\begin{document}

\title{NLH: A Blind Pixel-level Non-local Method \\
for Real-world Image Denoising}

\author{
  Yingkun Hou, \textit{Member, IEEE},
  Jun Xu,
  Mingxia Liu,
  Guanghai Liu, 
  Li Liu,
  Fan Zhu,
  Ling Shao
\thanks{
This work is supported by the National Natural Science Foundation of China (Grant No. 61379015, 61620106008 and 61866005).
YK Hou is with School of Information Science and Technology, Taishan University, Tai'an, China.
J Xu is with College of Computer Science, Nankai University, Tianjin, China.
MX Liu is with School of Medicine, The University of North Carolina at Chapel Hill, USA.
GH Liu is with School of Computer Science and Information Technology, Guangxi Normal University, China.
L Liu, F Zhu and L Shao are with Inception Institute of Artificial Intelligence and Mohamed bin Zayed University of Artificial Intelligence, Abu Dhabi, UAE.
J Xu (nankaimathxujun@gmail.com) is the corresponding author.
}
}


\maketitle

\begin{abstract}
Non-local self similarity (NSS) is a powerful prior of natural images for image denoising.\ Most of existing denoising methods employ similar patches, which is a patch-level NSS prior.\ In this paper, we take one step forward by introducing a pixel-level NSS prior, i.e., searching similar pixels across a non-local region.\ This is motivated by the fact that finding closely similar pixels is more feasible than similar patches in natural images, which can be used to enhance image denoising performance.\ With the introduced pixel-level NSS prior, we propose an accurate noise level estimation method, and then develop a blind image denoising method based on the lifting Haar transform and Wiener filtering techniques.\ Experiments on benchmark datasets demonstrate that, the proposed method achieves much better performance than previous non-deep methods, and is still competitive with existing state-of-the-art deep learning based methods on real-world image denoising.\ The code is publicly available at \url{https://github.com/njusthyk1972/NLH}.
\end{abstract}


%
\IEEEpeerreviewmaketitle

\section{Introduction}
\label{sec:intro}
Digital images are often subject to noise degradation during acquisition in imaging systems, due to the sensor characteristics and complex camera processing pipelines~\cite{crosschannel2016,see2018,burst2018}.\ Removing the noise from the acquired images is an indispensable step for image quality enhancement in low-level vision tasks~\cite{foi2008practical,Liang2018,STAR2020}.\ In general, image denoising aims to recover a clean image $\bm{x}$ from its noisy observation $\bm{y}=\bm{x}+\bm{n}$, where $\bm{n}$ is the corrupted noise.\ One popular assumption on $\bm{n}$ is additive white Gaussian noise (AWGN) with standard deviation (std) $\sigma$.\ Recently, increasing attention has been paid to removing realistic noise, which is more complex than AWGN~\cite{xuaccv2016,xu2018real,gid2018}. 

From the Bayesian perspective, image priors are of central importance for image denoising~\cite{lefkimmiatis2018,tian2020attention,tian2020image}.\ Numerous methods have been developed to exploit image priors for image denoising~\cite{pgpd,elad2006tip,ulyanov2018deep}
and other image restoration tasks~\cite{ren2019simultaneous,ren2019progressive,ren2018partial} over the past decades.\ These methods can be roughly divided into non-local self-similarity (NSS) based methods~\cite{nlm,bm3d,nlnet}, sparsity or low-rankness based methods~\cite{elad2006image,saist,wnnmijcv}, dictionary learning based methods \cite{ksvd,srcolor,lssc}, generative learning based methods~\cite{foe,epll,pajot2019}, and discriminative learning based methods \cite{mao2016image,dncnn,xu2019nac}, etc.\
\begin{figure}[t]
\vspace{-0mm}
\centering
\raisebox{-0.15cm}{\includegraphics[width=0.5\textwidth]{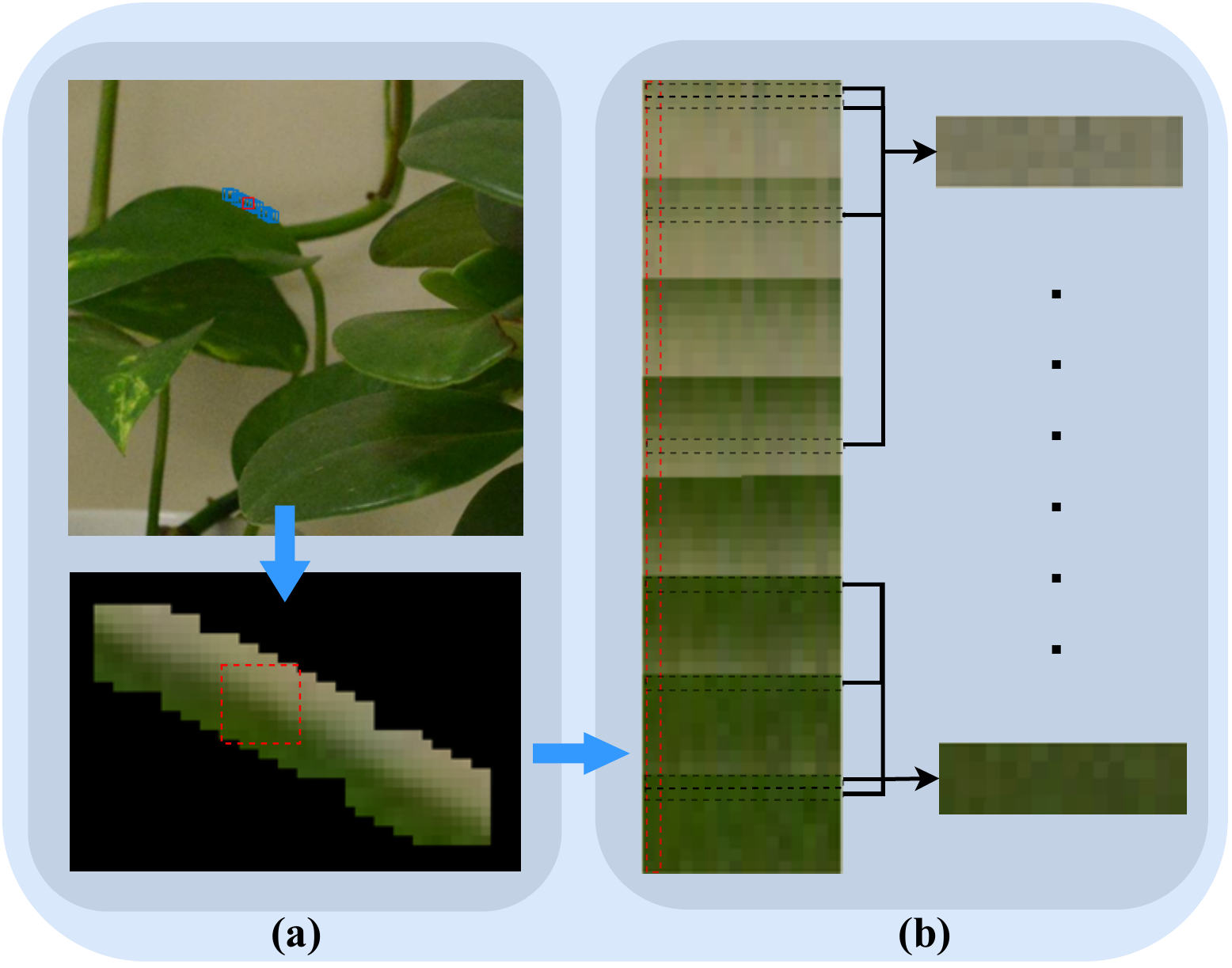}}
\vspace{-5mm}
\caption{\textbf{Illustration of the proposed searching scheme of non-local similar pixels}.\ (a) For each image patch (e.g., the patch in red box), we search its non-local similar patches (e.g., the patches in blue boxes).\ (b) Then we transform the similar patches into columns of vectors.\ For each row of pixels, we search their non-local similar pixels (e.g., the rows of pixels in black boxes) within the searched similar patches in (a).}
\vspace{-0mm}
\label{f-pixelnss}
\end{figure}

\begin{figure*}[ht!]
\centering
\begin{minipage}{0.12\textwidth}
\includegraphics[width=1\textwidth]{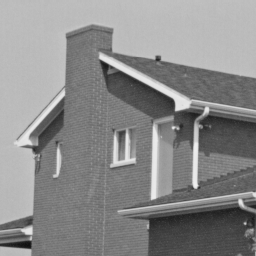}
\centering{\scriptsize (a) House}
\end{minipage}
\begin{minipage}{0.21\textwidth}
\vspace{-2mm}
\includegraphics[width=1\textwidth]{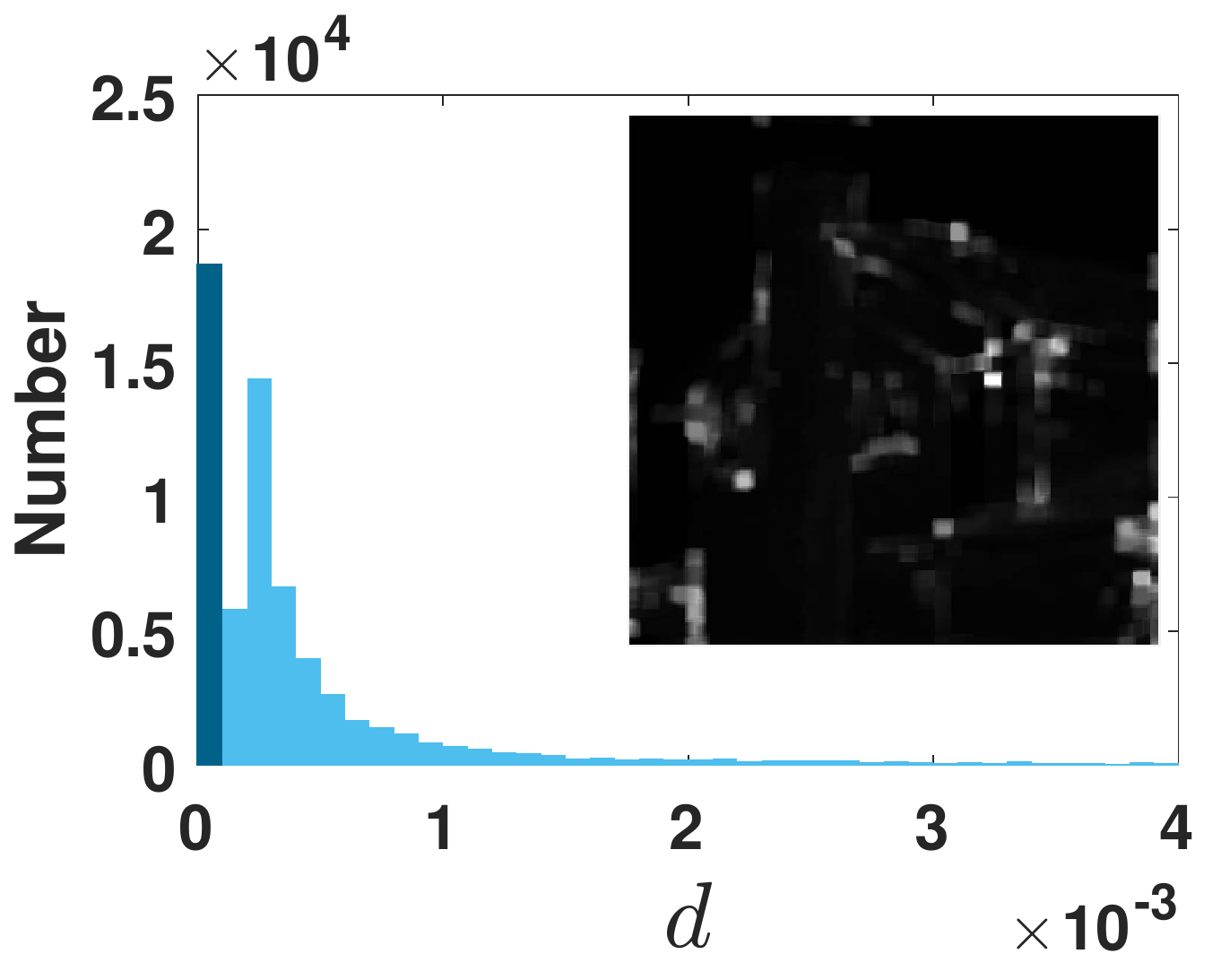}
\centering{\scriptsize (b) Patch Matching on Clean Image}
\end{minipage}
\begin{minipage}{0.21\textwidth}
\vspace{-2mm}
\includegraphics[width=1\textwidth]{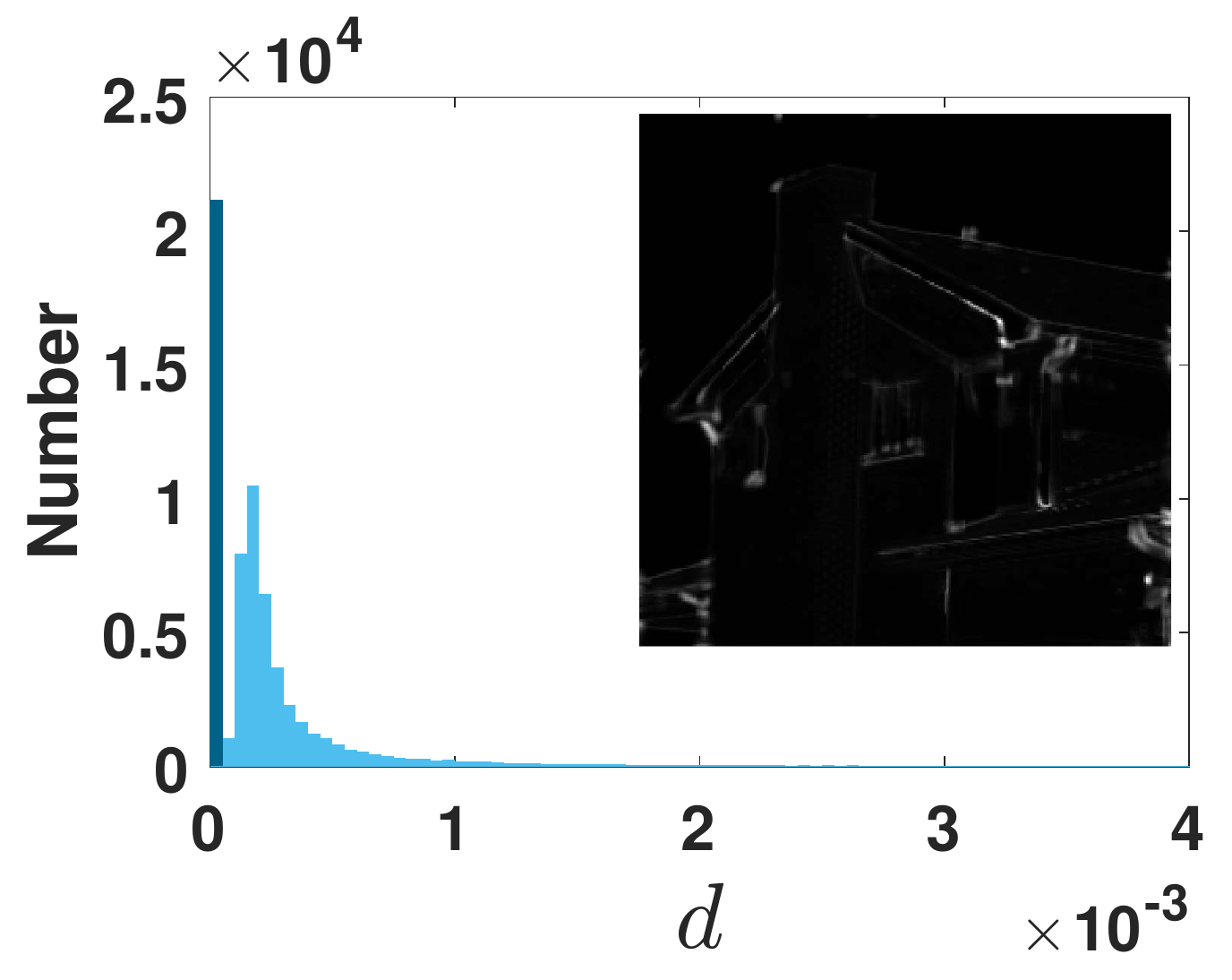}
\centering{\scriptsize (c) Pixel Matching on Clean Image}
\end{minipage}
\begin{minipage}{0.21\textwidth}
\includegraphics[width=1\textwidth]{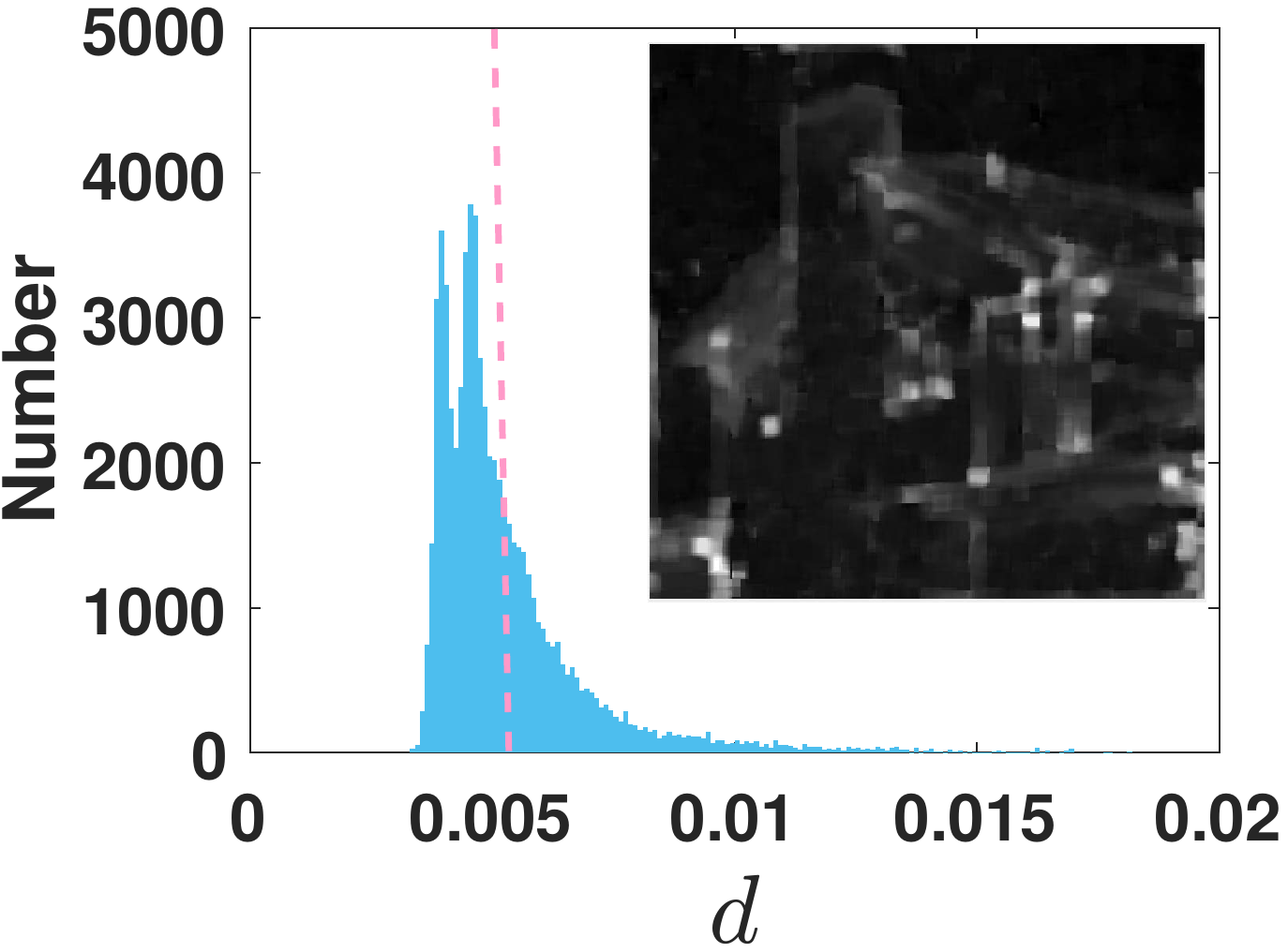}
\centering{\scriptsize (d) Patch Matching on Noisy Image}
\end{minipage}
\begin{minipage}{0.21\textwidth}
\includegraphics[width=1\textwidth]{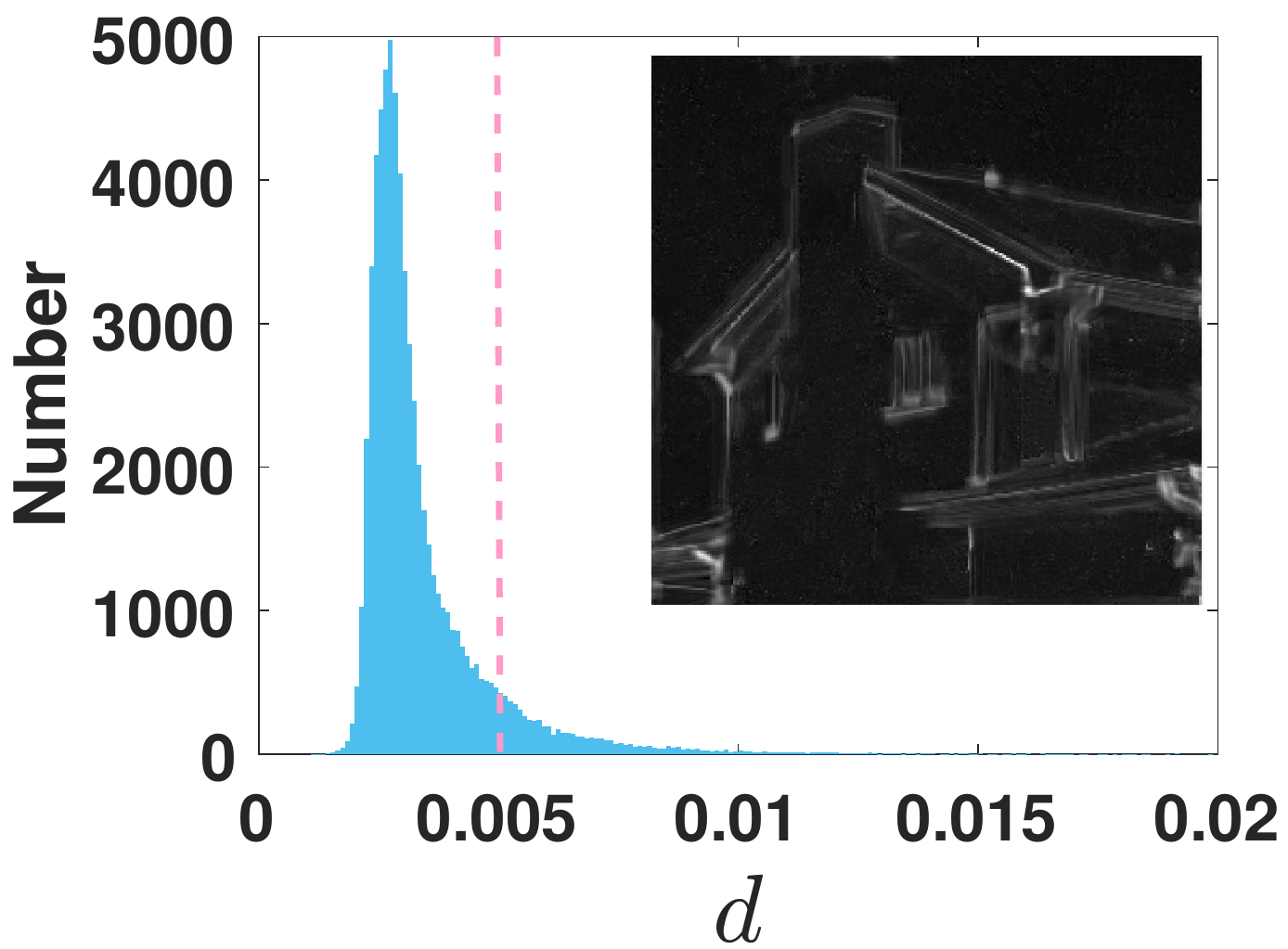}
\centering{\scriptsize (e) Pixel Matching on Noisy Image}
\end{minipage}
\vspace{-1mm}
\caption{\textbf{Histograms of the pixel-wise distance $d$ and the number of reference patches (or pixels)} whose pixel-wise distances to their corresponding most similar patches (or pixels) are $d$.\ Noisy image is generated by adding AWGN noise with $\sigma=15$ to (a).\ The images are normalized to $[0, 1]$. The pixel-wise distance maps are also plotted on the top-right corners of the corresponding histograms.
The darker bar in (b) (or (c)) highlights that the similar patches (or pixels) are with nearly zero pixel-wise distances to their corresponding reference patch (or reference row).
Comparing (d) and (e), we observe that in row matching, most of the pixel-wise distances by row matching (e) are less than 0.005 (dashed red vertical line).
However, in patch matching, much more patches suffer from the pixel-wise distances larger than 0.005 (dashed red vertical line). 
Comparing (d)-(e) and (b)-(c), we find that both patch and row matching in noisy images (d) and (e) have larger pixel-wise distances than those in corresponding clean images (b) and (c).}
\vspace{-2mm}
\label{f1}
\end{figure*}

Among the above-mentioned methods, the NSS prior arises from the fact that, in a natural image, a local patch has many non-local similar patches across the image.\ Here, the similarity is often measured by Euclidean distance.\ The NSS prior has been successfully utilized by state-of-the-art image denoising methods, such as BM3D \cite{bm3d,cbm3d,bm3dsapca}, WNNM \cite{wnnmijcv}, and N3Net \cite{n3net}, etc.\ However, most existing NSS-based methods \cite{ncsr,saist,mswnnm} perform identical noise removal on similar but nuanced patches, which would results in artifacts.\ Despite its capability to enhance denoising performance, this patch-level NSS prior employed in these methods suffers from one major bottleneck.\ That is, it is very challenging to find closely similar patches for all the reference patches in a natural image, especially when the number of similar patches is large.\ To break through this bottleneck, the strategy of searching shape adaptive similar patches is proposed in BM3D-SAPCA \cite{bm3dsapca}.\ Other improvements can also be found in~\cite{hou2011}.\ However, this would introduce shape artifacts into the denoised image.\ Multi-scale techniques \cite{Irani2013separating} have been proposed to enhance similarity, but the details would be degraded in the coarse scale and fail to detect similar counterparts.

In this work, we propose a pixel-level NSS prior for image denoising.\ 
The main idea of our work is illustrated in Figure~\ref{f-pixelnss}.\ 
Our motivation is that, since pixel is the smallest component of natural images, by lifting from patch-level to pixel-level, the NSS prior can be utilized to a greater extent.\ 
We evaluate this point through an example on the commonly used ``House'' image (Figure~\ref{f1} (a)).\ 
For each reference patch of size $8\times8$ in ``House'', we search its 16 most similar patches (including the reference patch itself) in the image by selecting the 16 minimum Euclidean distances which are computed between the reference patch and each patch in a $39\times39$ neighborhood. 
We then compute the average value of these 16 Euclidean distances and further divide this average value by the number of pixels (i.e., 64) to obtain a pixel-wise distance $d$.\ 
Here, the pixel-wise distance is defined as the distance apportioned to each pixel in this similar patch group which includes reference patch.\ 
We can see from Figures~\ref{f1} (b) and (d) that the distance maps are all block like by the patch-level self-similarity measurement.\
In Figure~\ref{f1} (b), we draw a histogram to show the relationship between the pixel-wise distance $d$ and the number of reference patches with given pixel-wise distance $d$ to their corresponding most similar patches.\ 
We observe that, less than $1.8\times10^{4}$ reference patches (the darker bar) closely match their corresponding similar patches.\ 

Then, we reshape each image patch in the similar patch group to 16 column vectors by column scanning and then stack them to form a $64\times16$ matrix, further implement row matching also by Euclidean distance between a reference row and each of other rows in this matrix, in each row matching, we select 4 most similar rows (including the reference row itself) to form a similar pixel matrix, we average these 4 distances and further divide it by 64 to obtain the pixel-wise distance.\
We plot the histogram in Figure~\ref{f1} (c) and one can see that the distance map value of each pixel by this method is different from others.\ 
We observe that, over $2.1\times10^{4}$ reference patches contain closely matched pixels. 
Since Figure~\ref{f1} (b) is the distance map between image patches, while Figure~\ref{f1} (c) is the distance map between different rows of pixels.\ 
Since each set of similar patches contains 64 groups of similar pixel rows, the number of pixel distance maps are 64 times of the patch distance maps.\ 
Therefore, for the same image, the bin size in patch distance map (b) should be larger than that in the pixel distance map (c).\ 
We then add AWGN noise ($\sigma=15$) to Figure~\ref{f1} (a), compute the pixel-wise distances in patch-level NSS (as (b)) and pixel-level NSS (as (c)), and draw the histograms in Figures.~\ref{f1} (d) and (e), respectively.\ We observe that, the histogram in Figure~\ref{f1} (e) is shifted to left with a large margin, when compared to that in Figure\ \ref{f1} (d).\ 
All these results demonstrate that, the proposed pixel-level NSS can exploit the capability of NSS prior to a greater extent than previous patch-level NSS.
\begin{figure*}[ht!]
\vspace{-0mm}
\centering
\raisebox{-0.15cm}{\includegraphics[width=1\textwidth]{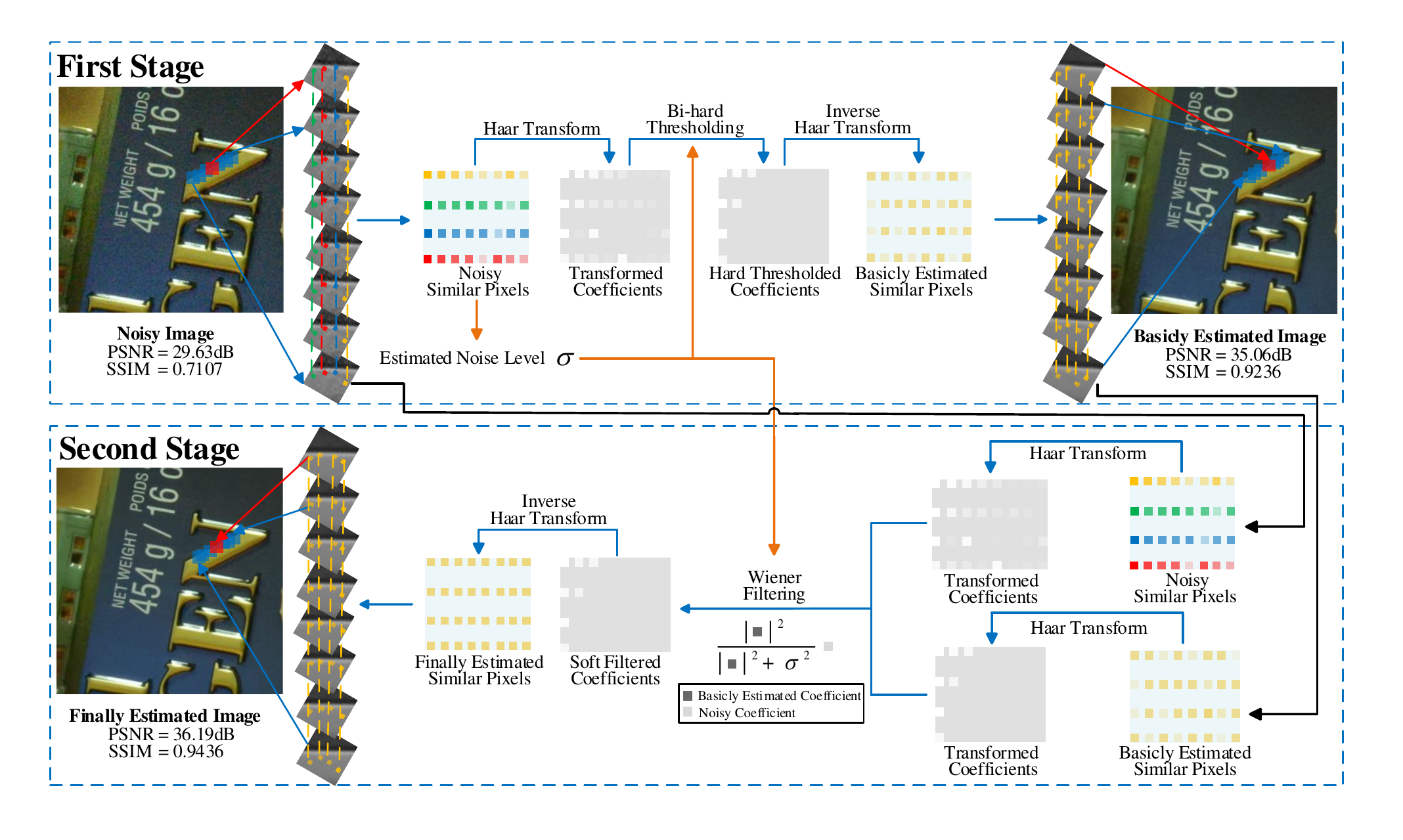}}
\vspace{-10mm}
\caption{\textbf{The overall framework of the proposed NLH method for real-world image denoising}.\ 
Our NLH consists of two stages.\ 
In the first stage, we extract non-local similar patches, and search similar pixels within these patches.\ 
The noise level is estimated on each group of similar pixels.\ 
For the matrix of similar pixels, we apply the Haar transform and bi-hard thresholding on it, and recover the basicly estimated pixels by inverse Haar transform.\ 
In the second stage, we utilize Haar transform on the similar pixel matrices of the original noisy image and basicly estimated image.\ 
The transformed coeficient matrices are then used to obtain a softly filtered coefficient matrix by Wiener filtering.\ 
Finally, we perform inverse Haar transform and get the finally estimated similar pixel matrices.\ 
These matrices of similar pixels are put together to form the finally estimated image.}
\vspace{-0mm}
\label{f2}
\end{figure*}

With the proposed pixel-level NSS prior, we develop an accurate noise level estimation method, and then propose a blind image denoising method based on non-local Haar (NLH) transform and Wiener filtering techniques.\ Experiments results show that, the proposed NLH method achieves much better performance than previous hand-crafted and non-deep methods, and is still competitive with existing state-of-the-art deep learning based methods on commonly tested real-world datasets.\ In summary, our contributions are manifold:
\begin{itemize}
\vspace{-0mm}
    \item We introduce a pixel-level NSS prior for image denoising, in which we find similar pixels instead of patches.
    \vspace{-0mm}
    \item With the pixel-level NSS prior, we propose an accurate noise level estimation method.\ Based on this, we propose a blind pixel-level image denoising method, and extend it for real-world image denoising.
    \vspace{-0mm}
    \item Extensive experiments on benchmark datasets demonstrate that, the proposed method achieves much better performance than previous non-deep methods, and is still competitive with existing state-of-the-art deep learning based methods on real-world image denoising.
    \vspace{-0mm}
\end{itemize}

The remainder of this paper is organized as follows.\ In \S\ref{sec:related}, we briefly survey the related work.\ In \S\ref{sec:method}, we present the proposed blind NLH method for image denoising.\ Extensive experiments are conducted in \S\ref{sec:exp} to evaluate its noise level estimation performance, and compare it with state-of-the-art image denoising methods on both synthetic and realistic noise removal.\ Conclusion is given in \S\ref{conclusion}.

\section{Related Work}
\label{sec:related}

\noindent
\textbf{Non-local Self Similarity (NSS)}: The NSS image prior is the essence to the success in texture synthesis \cite{efros1999texture}, image denoising \cite{bm3d}, super-resolution \cite{Irani2009sr}, and inpainting \cite{patchmatch}.\ In the domain of image denoising, the NSS prior is firstly employed by the Non-local Means (NLM) method \cite{nlm}.\ NLM estimates each pixel by computing a weighted average of all pixels in the image, where the weights are determined by the similarity between corresponding image patches centered at these pixels.\ Though this is a pixel-level method, NLM performs denoising based on the patch-level NSS prior.\ The patch-level NSS prior is l ater flourished in the BM3D method \cite{bm3d}, and also in \cite{pgpd,mcwnnm,twsc}.\ This prior performs denoising on groups of similar patches searched in non-local regions.\ These methods usually assume that the collected similar patches are fully matched.\ However, it is challenging to find closely similar patches for all the reference patches in a natural image.\ For more related work, please refer to~\cite{tian2019deep}.\ In this work, instead of only searching similar patches, we propose to further search similar pixels and perform pixel-level denoising accordingly.

\noindent
\textbf{Real-world Image Denoising}:\ Many real-world image denoising methods have been developed in the past decade \cite{crosschannel2016,mcwnnm,twsc}.\ The CBM3D method \cite{cbm3d} first transforms an input RGB image into the luminance-chrominance space (e.g., YCbCr) and then applies the BM3D method \cite{bm3d} to each channel separately.\ The method of \cite{liu2008automatic} introduces a ``noise level function" to estimate the noise of the input image and then removes the noise accordingly.\ The methods of \cite{noiseclinic,Zhu_2016_CVPR} perform blind image denoising by estimating the noise level in image patches.\ The method of \cite{crosschannel2016} employs a multivariate Gaussian to fit the noise in a noisy image and performs denoising accordingly.\ Neat Image \cite{neatimage} is a commercial software that removes noise according to the noise parameters estimated in a large enough flat region.\ MCWNNM \cite{mcwnnm} is a patch-level NSS prior based method, demanding a large number of similar patches for low-rank approximation.\ GCBD \cite{gcbd} is a blind image denoising method that uses the Generative Adversarial Network \cite{gan}.\ TWSC \cite{twsc} introduces a weighting scheme into the sparse coding model~\cite{pgpd} for real-world image denoising.\ It requires many similar patches for accurate weight calculation and denoising.\ Almost all these methods identically remove the noise in similar patches but ignore their internal variance.\ Besides, since the realistic noise in real-world images is pixel-dependent \cite{crosschannel2016,dnd2017,sidd2018}, patch-level NSS operations would generate artifacts when treating all the pixels alike.\ As such, real-world image denoising remains a very challenging problem \cite{dnd2017,sidd2018,xu2018real}. 

\section{Proposed Blind Pixel-level Denoising Method}
\label{sec:method}

In this section, we present the proposed pixel-level Non-local Haar transform (NLH) based method for blind image denoising.\ The overall method includes three parts: 1) searching non-local similar pixels (\S\ref{sec:pixel}), 2) noise level estimation (\S\ref{sec:level}), and 3) a two-stage framework for image denoising (\S\ref{sec:twostage}).\ The overall denoising framework is summarized in Figure\ \ref{f2}.\ In the first stage, we employ the lifting Haar transform~\cite{Sweldens1996,Sweldens1998} and bi-hard thresholding for local signal intensity estimation, which is later combined with the global noise level estimation for image denoising using Wiener filtering in the second stage.\ We then extend the proposed NLH method for real-world image denoising.

\subsection{Searching Non-local Similar Pixels}
\label{sec:pixel}
Given a gray-scale noisy image $\bm{y}\in\mathbb{R}^{h\times w}$, we extract its local patches (assume there are totally $N$ patches).\ We stretch each local patch of size $\sqrt{n}\times\sqrt{n}$ to a vector, denoted by $\bm{y}_{l,1}\in \mathbb{R}^{n}$ ($l=1,...,N$).\ For each $\bm{y}_{l,1}$, we search its $m$ most similar patches (including $\bm{y}_{l,1}$ itself) by Euclidean distance in a large enough window (of size $W\times W$) around it.\ We stack these vectors column by column (the patch with smaller distance is closer to the reference patch $\bm{y}_{l,1}$) to form a noisy patch matrix $\bm{Y}_l=[\bm{y}_{l,1},...,\bm{y}_{l,m}]\in\mathbb{R}^{n\times m}$.

To apply the NSS prior at the pixel-level, we further search similar pixels in $\bm{Y}_l$ by computing the Euclidean distances among the $n$ rows.\ Each row of $\bm{Y}_l$ contains $m$ pixels in the same relative position of different patches.\ The patch-level NSS prior guarantees that the pixels in the same row are similar to some extent.\ However, for rare textures and details, some pixels would suffer from large variance due to shape shifts.\ Processing these pixels identically would generate artifacts.\ To resolve this problem, we carefully select the pixels that are most similar to each other.\ Specifically, for the $i$-th row $\bm{y}_l^{i}\in\mathbb{R}^{m}$ of $\bm{Y}_l$, we compute the distance between it and the $j$-th row $\bm{y}_l^{j}$ ($j=1,..., n$) as
\vspace{-0mm}
\begin{equation}
\label{e1}
d_{l}^{ij}
=
\|
\bm{y}_l^{i}
-
\bm{y}_l^{j}
\|
_{2}
.
\vspace{-0mm}
\end{equation}
Note that $d_{l}^{ii}=0$ for each row $\bm{y}_l^{i}$.\ We then select the $q$ ($q$ is a power of $2$) rows, i.e., $\{\bm{y}_l^{i_1},...,\bm{y}_l^{i_q}\}$ ($i_1=i$, the row with smaller distance is closer to the reference row $\bm{y}_l^{i_1}$), in $\bm{Y}_l$ with the smallest distances to $\bm{y}_l^{i}$, and finally aggregate the similar pixel rows as a matrix $\bm{Y}_{l}^{iq}\in\mathbb{R}^{q\times m}$:
\vspace{-0mm}
\begin{equation}
\label{e2}
\bm{Y}_{l}^{iq}
=
\left[
 \begin{matrix}
   y_l^{i_1,1} & \cdots & y_l^{i_1,m} \\
  \vdots & \ddots & \vdots \\
   y_l^{i_q,1} & \cdots & y_l^{i_q,m} 
  \end{matrix}
  \right], 
  \vspace{-0mm}
\end{equation}
where $\{i_1,...,i_q\}$$\subset$$\{1,...,n\}$.\ The noisy pixel matrices $\{\bm{Y}_{l}^{iq}\}$ ($i=1,...,n; l=1,...,N$) in the whole image are used for noise level estimation, which is described as follows.

\subsection{Noise Level Estimation}
\label{sec:level}
Accurate and fast estimation of noise levels is an essential step for efficient image denoising.\ The introduced pixel-level NSS prior can help achieve this goal.\ The rationale is that, since the pixels in the selected $q$ rows of $\bm{Y}_{l}^{iq}$ are very similar to each other, the standard deviation (std) of among them can be viewed as the noise level.\ For simplicity, we assume that the noise follows a Gaussian distribution with std $\sigma_l$.\ Since the distances between the $i$-th row of $\bm{Y}_l$ and its most similar $q$ rows are $d_{l}^{ii_1},...,d_{l}^{ii_q}$ ($i_1=i$), $\sigma_l$ can be computed as 
\vspace{-0mm}
\begin{equation}
\label{e3}
\sigma_{l}
=
\frac{1}{n(q-1)}\sum_{t=2}^{q}\sum_{i=1}^{n}\sqrt{\frac{1}{m}(d_{l}^{ii_t})^2}
.
\vspace{-0mm}
\end{equation}

Initial experiments indicate that the Eqn.\ (\ref{e3}) performs well for smooth areas, but is problematic for textures and structures.\ This is because, in these areas, the signal and noise are difficult to distinguish, and thus the noise level would be over-estimated.\ To make our method more robust for noise level estimation, we extend the noise level estimation from a local region to a global one.\ To do so, we estimate the local noise levels for all the noisy pixel matrices in the image, and simply set the global noise level as
\vspace{-0mm}
\begin{equation}
\label{e4}
\sigma_{g}
=
\frac{1}{N}\sum_{l=1}^{N}\sigma_{l}
.
\vspace{-0mm}
\end{equation}

\textbf{Discussion}.\ The proposed pixel-based noise level estimation method assumes the noise in the selected $q$ rows follows a Gaussian distribution, which is consistent with the assumptions in \cite{crosschannel2016,twsc}.\ The proposed method is very simple, since it only computes the distances among the most similar pixels extracted from the image.\ As will be shown in the experimental section (\S\ref{sec:exp}), the proposed noise level estimation method is very accurate, which makes it feasible to develop a blind image denoising method for real-world applications.\ Now we introduce the proposed two-stage denoising framework below.

\subsection{Two-stage Denoising Framework}
\label{sec:twostage}
The proposed denoising method consists of two stages. In the first stage, we estimate the local intensities via the non-local Haar (NLH) transform based bi-hard thresholding.\ With the results from the first stage, we perform blind image denoising by employing Wiener filtering based soft thresholding, in the second stage.\ Now, we introduce the two stages in details.

\textbf{Stage\ 1:\ Local Intensity Estimation by Lifting Haar Transform based Bi-hard Thresholding}.\ We have grouped a set of matrices $\bm{Y}_l^{q}\in\mathbb{R}^{q\times m}$ ($l=1,...,N$) consisting of similar pixels.\ For simplicity, we ignore the index $i$) and estimate the global noise level $\sigma_g$.\ We perform denoising on the matrices consisting of similar pixels in the Haar transformed domain \cite{Haar1910}.\ Here, we utilize the lifting Haar wavelet transform (LHWT)~\cite{Sweldens1996,Sweldens1998} due to its flexible operation, faster speed, and lightweight memory.\

The LHWT matrices we employ here are two orthogonal matrices $\bm{H}_{l}\in\mathbb{R}^{q\times q}$ and $\bm{H}_{r}\in\mathbb{R}^{m\times m}$.\ We set $q, m$ as powers of $2$ to accommodate the noisy pixel matrices $\{\bm{Y}_l^{q}\}_{l=1}^{N}$ with the Haar transform.\ The LHWT transform of the non-local similar pixel matrix $\bm{Y}_l^{q}$, i.e., the matrix consist of similar pixels is to obtain the transformed noisy coefficient matrix $\bm{C}_{l}^{q}\in\mathbb{R}^{q\times m}$ via
\vspace{-0mm}
\begin{equation}
\label{e5}
\bm{C}_{l}^{q}
=
\bm{H}_{l}\bm{Y}_l^{q}\bm{H}_{r}
.
\vspace{-0mm}
\end{equation}
Due to limited space, we put the detailed LHWT transforms with specific $\{q,m\}$ in the~\emph{Appendix}.

After LHWT transforms, we restore the $j$-th ($j=1,...,m$) element in $i$-th row ($i=1,...,q$) of the noisy coefficient matrix $\bm{C}_{l}^{q}$ via hard thresholding:
\vspace{-0mm}
\begin{equation}
\label{e6}
\bm{\hat{C}}_{l}^{q}
=
\bm{C}_{l}^{q}
\odot
\mathbb{I}_{
\left\{
|\bm{C}_{l}^{q}|\ge\tau\sigma_{g}^{2}
\right\}
}
,
\vspace{-0mm}
\end{equation}
where $\odot$ means element-wise production, $\mathbb{I}$ is the indicator function, and $\tau$ is the threshold parameter.\ According to the wavelet theory \cite{Sweldens1996}, the coefficients in the last two rows of $\bm{C}_{l}^{q}$ (except the $1$-st column) are in the high frequency bands of the LHWT transform, which should largely be noise.\ To remove this noise in $\bm{C}_{l}^{q}$, we introduce a structurally hard thresholding strategy and completely set to $0$ all the coefficients in the high frequency bands of $\bm{\hat{C}}_{l}^{q}$:
\vspace{-0mm}
\begin{equation}
\label{e7}
\bm{\widetilde{C}}_{l}^{q}(i,j)
=
\bm{\hat{C}}_{l}^{q}(i,j)
\odot
\mathbb{I}_{
\left\{
\text{if}\ i=1,...,q-2\ \text{or}\ j=1
\right\}
}
,
\vspace{-0mm}
\end{equation}
where $\bm{\widetilde{C}}_{l}^{q}(i,j)$ and $\bm{\hat{C}}_{l}^{q}(i,j)$ are the $i,j$-th entry of the coefficient matrices $\bm{\widetilde{C}}_{l}^{q}$ and $\bm{\hat{C}}_{l}^{q}$, respectively.\ We then employ inverse LHWT transforms~\cite{Sweldens1996,Sweldens1998} on $\bm{\widetilde{C}}_{l}^{q}$ to obtain the denoised pixel matrix $\bm{\widetilde{Y}}_{l}^{q}$ via
\vspace{-0mm}
\begin{equation}
\label{e8}
\bm{\widetilde{Y}}_{l}^{q}
=
\bm{H}_{il}
\bm{\widetilde{C}}_{l}^{q}(i,j)
\bm{H}_{ir}
,
\vspace{-0mm}
\end{equation}
where $\bm{H}_{il}\in\mathbb{R}^{q\times q}$ and $\bm{H}_{ir}\in\mathbb{R}^{m\times m}$ are inverse LHWT matrices.\ Detailed inverse LHWT with specific $\{q,m\}$ are put in the~\emph{Appendix}.\
Finally, we aggregate all the denoised pixel matrices to form the denoised image. 
The elements in $\bm{\widetilde{C}}_{l}^{q}$ can be viewed as local signal intensities, which are used in \textbf{Stage 2} for precise denoising with the globally estimated noise level $\sigma_g$.\ To obtain more accurate estimation of local signal intensities, we perform the above LHWT transform based bi-hard thresholding for $K$ iterations.\ For the $k$-th ($k=1,...,K$) iteration, we add the denoised image $\bm{y}_{k-1}$ back to the original noisy image $\bm{y}$ and obtain the noisy image $\bm{y}_{k}$ as
\begin{equation}
\label{e9}
\bm{y}_{k}
=
\lambda
\bm{y}_{k-1}
+
(1-\lambda)
\bm{y}
.
\vspace{-0mm}
\end{equation}

\begin{table*}[t]
\centering
\small
\caption{Estimated noise levels of different methods on the \textbf{BSD68} dataset corrupted by AWGN noise with std $\sigma$. ``-'' indicates that the results cannot be obtained due to the internal errors of the code.}
\vspace{-3mm}
\begin{center}
\begin{tabular}{r||ccccccc}
\Xhline{1pt}
\rowcolor[rgb]{ .85,  .9,  0.95}
Noise std $\sigma$
& 5 & 15 & 25 & 35 & 50 & 75 & 100
\\
\hline
Zoran \textsl{et al.} \cite{zoran2009scale} 
& 4.74 & 14.42 & - & - & 49.23 & 74.33 & - 
\\
Liu \textsl{et al.} \cite{noiselevel} 
& \textbf{5.23} & \textbf{15.18} & \textbf{25.13} & \textbf{34.83} & 49.54 & 74.36 & 98.95 
\\
Chen \textsl{et al.}\ ~\cite{Chen2015ICCV}
& 8.66 & 16.78 & 26.26 & 36.00 & 50.82 & 75.75 & 101.62 
\\
\hline
\rowcolor[rgb]{ .85,  .85,  .85}
Our Method (Eqn.\ (\ref{e4}))
& 5.91 & 15.88 & 25.64 & 35.50 & \textbf{50.45} & \textbf{75.40} & \textbf{100.97} 
\\
\hline
\end{tabular}
\end{center}
\vspace{-1mm}
\label{t1}
\vspace{-0mm}
\end{table*}

\textbf{Stage\ 2:\ Blind Denoising by Iterative Wiener Filtering}.\ 
%
The denoised image in the \textbf{Stage 1} is only a basic estimation of the latent clean image, and it is taken as a reference image for further fine-grained denoising in this stage via Wiener filtering.\ In order to remove the noise more clear while preserving the details, we employ the Wiener filtering based soft thresholding for finer denoising.\ We use the above estimated local signal intensities and the globally estimated noise level $\sigma_{g}$ to perform Wiener filtering on the coefficients obtained by the LHWT transform of the original noisy pixel matrices.\ To further improve the denoising performance, in all experiments, we conduct the Wiener filtering based soft thresholding for two iterations.\ In the first iteration, we perform Wiener filtering on $\bm{C}_{l}^{q}$ in Eqn.\ (\ref{e5}) as
\vspace{-1mm}
\begin{equation}
\label{e10}
\overline{\bm{C}_{l}^{q}}(i,j)
=
\frac{|\bm{\widetilde{C}}_{l}^{q}(i,j)|^{2}}
{|\bm{\widetilde{C}}_{l}^{q}(i,j)|^{2}+(\sigma_{g}/2)^{2}}
\bm{C}_{l}^{q}(i,j)
,
\vspace{-1mm}
\end{equation}
and then we perform the second Wiener filtering as
\vspace{-0mm}
\begin{equation}
\label{e11}
\overline{\overline{\bm{C}_{l}^{q}}}(i,j)
=
\frac{|\bm{\widetilde{C}}_{l}^{q}(i,j)|^{2}}
{|\bm{\widetilde{C}}_{l}^{q}(i,j)|^{2}+(\sigma_{g}/2)^{2}}
\overline{\bm{C}_{l}^{q}}(i,j)
.
\vspace{-0mm}
\end{equation}

Note that both the original noisy image and the roughly denoised image are required in the \textbf{Stage 2}.\ 
Just as shown by Eqns.\ (\ref{e10}) and (\ref{e11}), to perform Wiener filtering, we need the Haar transformed coefficients of both the original noisy image and the roughly denoised image in the \textbf{Stage 2}.\ To this end, we need simultaneously transform the matrix of similar pixels from the original noisy image, and that of the denoised image obtained in the \textbf{Stage 1}, to implement the Wiener filtering in the \textbf{Stage 2} of our NLH.\
Experiments on image denoising demonstrate that, the proposed method with two iterations performs the best, while using more iterations brings little improvement.\ 
We then perform inverse LHWT transforms (please see details in the~\emph{Appendix}) on $\overline{\overline{\bm{C}_{l}^{q}}}$ to obtain the denoised pixel matrix $\overline{\overline{\bm{Y}_{l}^{q}}}$.\ 
Finally, we aggregate all the denoised pixel matrices to form the final denoised image.\

\subsection{Complexity Analysis}
\label{sec:complexity}
The proposed NLH contains three parts:\ 1) In \S\ref{sec:pixel}, the complexity of searching similar patches is $\mathcal{O}(NW^2n)$, while the complexity of searching similar pixels is $\mathcal{O}(Nn^2m)$.\ Since we set $W>n>m$, the overall complexity is $\mathcal{O}(NW^2n)$.\ 2) In \S\ref{sec:level}, the complexity for noise level estimation is $\mathcal{O}(Nnq)$, which can be ignored.\ 3) In \S\ref{sec:twostage}, the complexity of the two stages are $\mathcal{O}(KNnm)$ and $\mathcal{O}(Nnm)$, respectively.\ Since we have $m>K$, the complexity of NLH is $\mathcal{O}(NW^2n)$.

\subsection{Extension to Real-world Image Denoising}
\label{sec:extend}
To accommodate the proposed NLH method with real-world RGB images, we first transform the RGB images into the luminance-chrominance (e.g., YCbCr) space \cite{bm3d}, and then perform similar pixel searching in the Y channel.\ The similar pixels in the other two channels (i.e., Cb and Cr) are correspondingly grouped.\ We perform denoising for each channel separately and aggregate the denoised channels back to form the denoised YCbCr image.\ Finally, we transform it back to the RGB space for visualization.

\section{Experiments and Results}
\label{sec:exp}
In this section, we first evaluate the developed noise level estimation method on synthetic noisy images.\ The goal is to validate the effectiveness of our pixel-level non-local self similarity (NSS) prior.\ We then evaluate the proposed NLH method on both synthetic images corrupted by additive white Gaussian noise (AWGN) and real-world noisy images.\ Finally, we perform comprehensive ablation studies to gain a deeper insight into the proposed NLH method.\ More results on visual quality can be found in the~\emph{Supplementary File}.
\begin{table*}[t!]
\vspace{-0mm}
\caption{Average PSNR(dB)/SSIM results of different methods on 20 gray-scale images corrupted by AWGN noise.}
\vspace{-3mm}
\begin{center}
\begin{tabular}{r||c|c|c|c|c|c|c|c|c|c}
\Xhline{1pt}
\rowcolor[rgb]{ .85,  .9,  .95}
\multicolumn{1}{c||}{Noise std $\sigma$}
&
\multicolumn{2}{c|}{15}
&
\multicolumn{2}{c|}{25}
&
\multicolumn{2}{c|}{35}
&
\multicolumn{2}{c|}{50}
&
\multicolumn{2}{c}{75}
\\
\rowcolor[rgb]{ .85,  .9,  .95}
\multicolumn{1}{c||}{Metric}
&PSNR$\uparrow$&SSIM$\uparrow$
&PSNR$\uparrow$&SSIM$\uparrow$
&PSNR$\uparrow$&SSIM$\uparrow$
&PSNR$\uparrow$&SSIM$\uparrow$
&PSNR$\uparrow$&SSIM$\uparrow$
\\
\hline
\hline
\textbf{NLM}\ ~\cite{nlm}
&31.20&0.8483&28.64&0.7602&26.82&0.6762&24.80&0.5646&22.43&0.4224 
\\
\textbf{BM3D}~\cite{bm3dsapca}
&32.42&0.8860&30.02&0.8364&28.48&0.7969&26.85&0.7481&24.74&0.6649 
\\
\textbf{LSSC}~\cite{lssc}
&32.27&0.8849&29.84&0.8329&28.26&0.7908&26.64&0.7405&24.77&0.6746
\\
\textbf{NCSR}~\cite{ncsr}
&32.19&0.8814&29.76&0.8293&28.17&0.7855&26.55&0.7391&24.66&0.6793
\\
\textbf{WNNM}~\cite{wnnmijcv}
&32.43&0.8841&30.05&0.8365&28.51&0.7958&26.92&0.7499&25.15&0.6903
\\
\textbf{TNRD}\ ~\cite{chen2017trainable}
&32.48&0.8845&30.07&0.8366&28.53&0.7957&26.95&0.7495&25.10&0.6901
\\
\textbf{DnCNN}~\cite{dncnn}
&32.59&0.8879&30.22&0.8415&28.66&0.8021&27.08&0.7563&25.24&0.6931
\\
\hline
\rowcolor[rgb]{ .85,  .85,  .85}
\textbf{NLH} (Blind)
&32.28&0.8796&30.09&0.8355&28.60&0.7988&27.11&0.7524&25.31&0.6932
\\
\hline
\end{tabular}
\vspace{-1mm}
\end{center}
\label{t2}
\vspace{-0mm}
\end{table*}

\vspace{-0mm}
\subsection{Implementation Details}
\label{sec:expdetail}
\vspace{-0mm}
The proposed NLH method has 7 main parameters: patch size $\sqrt{n}$, window size $W$ for searching similar patches, number of similar patches $m$, number of similar pixels $q$, regularization parameter $\lambda$, hard threshold parameter $\tau$, and iteration number $K$ ($\lambda$, $\tau$, $K$ only exist in \textbf{Stage 1}).\ In all experiments, we set $W=40$, $m=16$, $q=4$, $\tau=2$, $\lambda=0.6$.\ For synthetic AWGN corrupted image denoising, we set $\sqrt{n}=8, K=4$ for $0<\sigma\leq50$, $\sqrt{n}=10,K=5$ for $\sigma>50$ in both stages.\ For real-world image denoising, we set $\sqrt{n}=7$, $K=2$ in both stages.

\begin{figure*}[ht!]
\vspace{-2mm}
\centering
\subfigure{
\begin{minipage}{0.09\textwidth}
\includegraphics[width=1\textwidth]{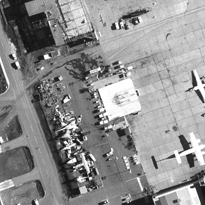}
\end{minipage}
\begin{minipage}{0.09\textwidth}
\includegraphics[width=1\textwidth]{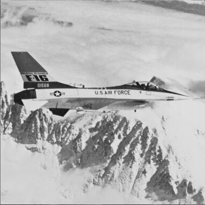}
\end{minipage}
\begin{minipage}{0.09\textwidth}
\includegraphics[width=1\textwidth]{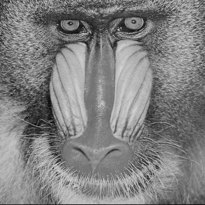}
\end{minipage}
\begin{minipage}{0.09\textwidth}
\includegraphics[width=1\textwidth]{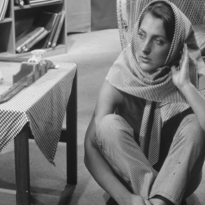}
\end{minipage}
\begin{minipage}{0.09\textwidth}
\includegraphics[width=1\textwidth]{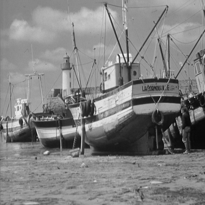}
\end{minipage}
\begin{minipage}{0.09\textwidth}
\includegraphics[width=1\textwidth]{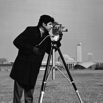}
\end{minipage}
\begin{minipage}{0.09\textwidth}
\includegraphics[width=1\textwidth]{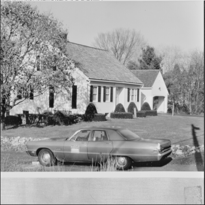}
\end{minipage}
\begin{minipage}{0.09\textwidth}
\includegraphics[width=1\textwidth]{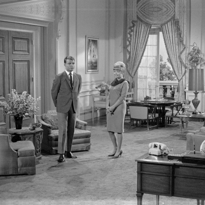}
\end{minipage}
\begin{minipage}{0.09\textwidth}
\includegraphics[width=1\textwidth]{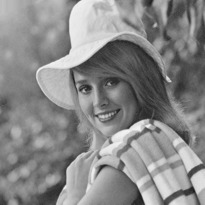}
\end{minipage}
\begin{minipage}{0.09\textwidth}
\includegraphics[width=1\textwidth]{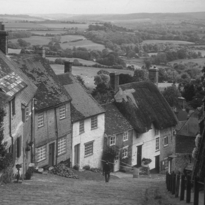}
\end{minipage}
}\vspace{-2mm}
\subfigure{
\begin{minipage}{0.09\textwidth}
\includegraphics[width=1\textwidth]{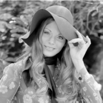}
\end{minipage}
\begin{minipage}{0.09\textwidth}
\includegraphics[width=1\textwidth]{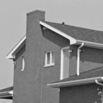}
\end{minipage}
\begin{minipage}{0.09\textwidth}
\includegraphics[width=1\textwidth]{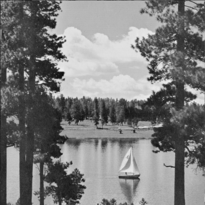}
\end{minipage}
\begin{minipage}{0.09\textwidth}
\includegraphics[width=1\textwidth]{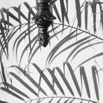}
\end{minipage}
\begin{minipage}{0.09\textwidth}
\includegraphics[width=1\textwidth]{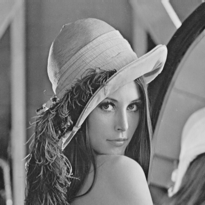}
\end{minipage}
\begin{minipage}{0.09\textwidth}
\includegraphics[width=1\textwidth]{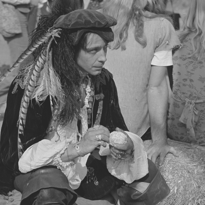}
\end{minipage}
\begin{minipage}{0.09\textwidth}
\includegraphics[width=1\textwidth]{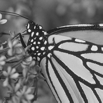}
\end{minipage}
\begin{minipage}{0.09\textwidth}
\includegraphics[width=1\textwidth]{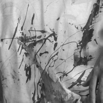}
\end{minipage}
\begin{minipage}{0.09\textwidth}
\includegraphics[width=1\textwidth]{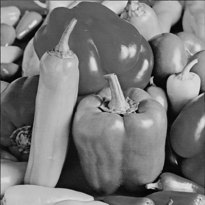}
\end{minipage}
\begin{minipage}{0.09\textwidth}
\includegraphics[width=1\textwidth]{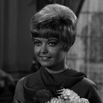}
\end{minipage}
}\vspace{-1mm}
\caption{The 20 commonly used gray-scale test images.}
\vspace{-2mm}
\label{f-20images}
\end{figure*}

\vspace{-1mm}
\subsection{Results on Noise Level Estimation}
\label{sec:explevel}
\vspace{-0mm}
The proposed pixel-level NSS prior can be used to estimate the noise level of the input noisy image.\ We compare our method (Eqn.\ (\ref{e4})) with leading noise level estimation methods, such as Zoran \textsl{et al.} \cite{zoran2009scale}, Liu \textsl{et al.} \cite{noiselevel}, and Chen \textsl{et al.} \cite{Chen2015ICCV}.\ The comparison is performed on the 68 images from the commonly tested \textbf{BSD68} dataset.\ We generate synthetic noisy images by adding AWGN with $\sigma$$\in$$\{5,15,25,35,50,75,100\}$ to the clean images.\ The comparison results are listed in Table \ref{t1}.\ The presented noise levels of different methods are averaged on the whole dataset. One can see that, the proposed method can accurately estimate different noise levels for various noisy images.\ Note that the proposed method only utilizes the introduced pixel-level NSS prior, and the results indeed validate its effectiveness on noise level estimation.\ 

\begin{figure*}[ht!]
\vspace{-0mm}
\centering
\subfigure{
\begin{minipage}{0.11\textwidth}
\includegraphics[width=1\textwidth]{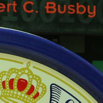}
\end{minipage}
\begin{minipage}{0.11\textwidth}
\includegraphics[width=1\textwidth]{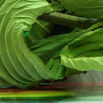}
\end{minipage}
\begin{minipage}{0.11\textwidth}
\includegraphics[width=1\textwidth]{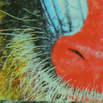}
\end{minipage}
\begin{minipage}{0.11\textwidth}
\includegraphics[width=1\textwidth]{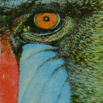}
\end{minipage}
\begin{minipage}{0.11\textwidth}
\includegraphics[width=1\textwidth]{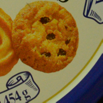}
\end{minipage}
\begin{minipage}{0.11\textwidth}
\includegraphics[width=1\textwidth]{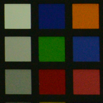}
\end{minipage}
\begin{minipage}{0.11\textwidth}
\includegraphics[width=1\textwidth]{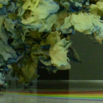}
\end{minipage}
\begin{minipage}{0.11\textwidth}
\includegraphics[width=1\textwidth]{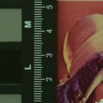}
\end{minipage}
}\vspace{-2mm}
\subfigure{
\begin{minipage}{0.11\textwidth}
\includegraphics[width=1\textwidth]{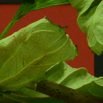}
\end{minipage}
\begin{minipage}{0.11\textwidth}
\includegraphics[width=1\textwidth]{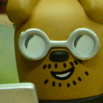}
\end{minipage}
\begin{minipage}{0.11\textwidth}
\includegraphics[width=1\textwidth]{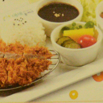}
\end{minipage}
\begin{minipage}{0.11\textwidth}
\includegraphics[width=1\textwidth]{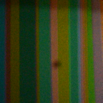}
\end{minipage}
\begin{minipage}{0.11\textwidth}
\includegraphics[width=1\textwidth]{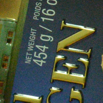}
\end{minipage}
\begin{minipage}{0.11\textwidth}
\includegraphics[width=1\textwidth]{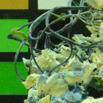}
\end{minipage}
\begin{minipage}{0.11\textwidth}
\includegraphics[width=1\textwidth]{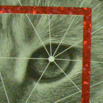}
\end{minipage}
}\vspace{-0mm}
\caption{The 15 cropped real-world noisy images from the \textbf{CC} dataset \cite{crosschannel2016}.}
\label{f-cc}
\end{figure*}

\subsection{\parbox[t]{10cm}{Results on Synthetic AWGN Corrupted Images}}
\label{sec:expawgn}

On 20 gray-scale images (listed in Figure\ \ref{f-20images}) widely used in \cite{bm3d,wnnmijcv,pgpd}, we compare the proposed NLH method with several competing AWGN denoising methods, such as BM3D \cite{bm3d}, LSSC \cite{lssc}, NCSR \cite{ncsr}, WNNM \cite{wnnmijcv}, TNRD \cite{chen2017trainable}, and DnCNN \cite{dncnn}.\ For BM3D, we employ its extension called BM3D-SAPCA \cite{bm3dsapca}, which usually performs better than BM3D on gray-scale images.\ We employ the Non-Local Means (NLM) \cite{nlm} as a baseline to validate the effectiveness of the pixel-level NSS prior.\ The source codes of these methods are downloaded from the corresponding authors' websites, and we use the default parameter settings.\ The methods of TNRD and DnCNN are discriminative learning based methods, and we use the models trained originally by their authors.\ The noisy image is generated by adding AWGN noise with standard deviation (std) $\sigma$ to the corresponding clean image, and in this paper we set $\sigma\in\{15, 25, 35, 50, 75\}$.\ Note that the noise level $\sigma$ is the same for each image of the whole dataset.

\begin{figure*}
\centering
\subfigure{
\begin{minipage}[t]{0.19\textwidth}
\centering
\raisebox{-0.15cm}{\includegraphics[width=1\textwidth]{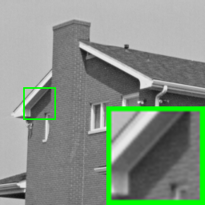}}
{\footnotesize (a) Ground Truth}
\end{minipage}
\begin{minipage}[t]{0.19\textwidth}
\centering
\raisebox{-0.15cm}{\includegraphics[width=1\textwidth]{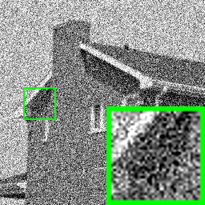}}
{\footnotesize (b) Noisy (14.12/0.1253)}
\end{minipage}
\begin{minipage}[t]{0.19\textwidth}
\centering
\raisebox{-0.15cm}{\includegraphics[width=1\textwidth]{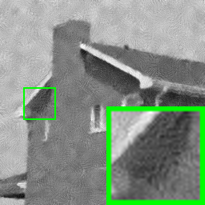}}
{\footnotesize (c) NLM (25.94/0.5513)}
\end{minipage}
\begin{minipage}[t]{0.19\textwidth}
\centering
\raisebox{-0.15cm}{\includegraphics[width=1\textwidth]{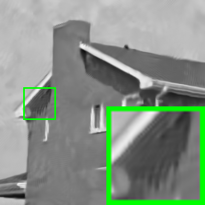}}
{\footnotesize (d) BM3D-S (29.50/0.8081)}
\end{minipage}
\begin{minipage}[t]{0.19\textwidth}
\centering
\raisebox{-0.15cm}{\includegraphics[width=1\textwidth]{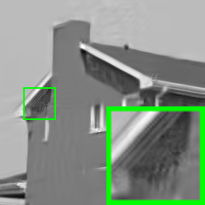}}
{\footnotesize (e) LSSC (29.97/0.8174)}
\end{minipage}
}\vspace{-3mm}
\subfigure{
\begin{minipage}[t]{0.19\textwidth}
\centering
\raisebox{-0.15cm}{\includegraphics[width=1\textwidth]{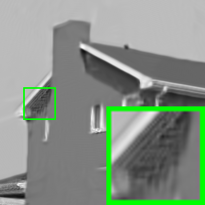}}
{\footnotesize (f) NCSR (29.62/0.8156) }
\end{minipage}
\begin{minipage}[t]{0.19\textwidth}
\centering
\raisebox{-0.15cm}{\includegraphics[width=1\textwidth]{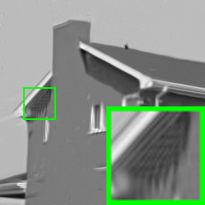}}
{\footnotesize (g) WNNM (30.31/0.8215)}
\end{minipage}
\begin{minipage}[t]{0.19\textwidth}
\centering
\raisebox{-0.15cm}{\includegraphics[width=1\textwidth]{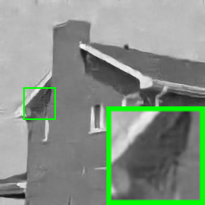}}
{\footnotesize (h) TNRD (29.48/0.8064)}
\end{minipage}
\begin{minipage}[t]{0.19\textwidth}
\centering
\raisebox{-0.15cm}{\includegraphics[width=1\textwidth]{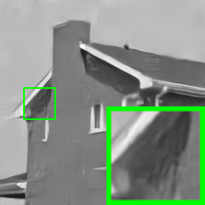}}
{\footnotesize (i) DnCNN (30.01/0.8192)}
\end{minipage}
\begin{minipage}[t]{0.19\textwidth}
\centering
\raisebox{-0.15cm}{\includegraphics[width=1\textwidth]{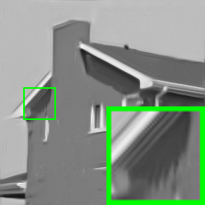}}
{\footnotesize (j) NLH (\textbf{30.50}/\textbf{0.8258}) }
\end{minipage}
}\vspace{-1mm}
\caption{\textbf{Denoised images and PSNR(dB)/SSIM results of \textsl{House} by different methods} (the noise level is $\sigma=50$).\ ``BM3D-S" is an abbreviation of ``BM3D-SAPCA" due to limited space.}
\label{f-awgn}
\end{figure*}

From Table~\ref{t2} we can see that, the proposed NLH is comparable with the leading denoising methods on average PSNR (dB) and SSIM~\cite{ssim}.\ Note that TNRD and DnCNN are trained on clean and synthetic noisy image pairs, while NLH can blindly remove the noise with the introduced pixel-level NSS prior.\ By comparing the performance of NLM and NLH, one can see that the proposed pixel-level denoising method performs much better than simply averaging the central pixels of similar patches.\ The visual quality comparisons can be found in Figure~\ref{f-awgn}.\ We observe that our NLH produces more visual pleasing results than the other methods.

\subsection{Results on Real-World Noisy Images}
\label{sec:expreal}
\begin{table*}
\vspace{-0mm}
\caption{PSNR(dB) results of different methods on the 15 cropped real-world noisy images in \textbf{CC} dataset~\cite{crosschannel2016}.}
\begin{center}
\renewcommand\arraystretch{1}
\footnotesize
\begin{tabular}{c||c|cccccccccc}
\Xhline{1pt}
\rowcolor[rgb]{ .85,  .9,  .95}
Camera Settings  
&
\#
&
\textbf{CBM3D}
&
\textbf{NI}
&
\textbf{NC}
&
\textbf{CC}
&
\textbf{\footnotesize MCWNNM} 
&
\textbf{TWSC}
&
\textbf{DnCNN+}
&
\textbf{FFDNet+}
&
\textbf{CBDNet}
&
\textbf{NLH}
\\
\hline
\multirow{3}{*}{\small{Canon 5D M3}}  
& 1 & 39.76 & 35.68 & 36.20 & 38.37 & 41.13 & 40.76 & 38.02 & 39.35 & 36.68 & \textbf{41.57}
\\
\multirow{3}{*}{ISO = 3200} 
& 2 & 36.40 & 34.03 & 34.35 & 35.37 & 37.28 & 36.02 & 35.87 & 36.99 & 35.58 & \textbf{37.39}
\\ 
& 3 & 36.37 & 32.63 & 33.10 & 34.91 & 36.52 & 34.99 & 35.51 & 36.50 & 35.27 & \textbf{36.68}
\\
\hline
\multirow{3}{*}{Nikon D600} 
& 4 & 34.18 & 31.78 & 32.28 & 34.98 & \textbf{35.53} & 35.32 & 34.75 & 34.96 & 34.01 & 35.50
\\ 
\multirow{3}{*}{ISO = 3200}   
& 5 & 35.07 & 35.16 & 35.34 & 35.95 & 37.02 & 37.10 & 35.28 & 36.70 & 35.19 & \textbf{37.21}
\\ 
& 6 & 37.13 & 39.98 & 40.51 & 41.15 & 39.56 & 40.90 & 37.43 & 40.94 & 39.80 & \textbf{41.34}
\\
\hline
\multirow{3}{*}{Nikon D800} 
& 7 & 36.81 & 34.84 & 35.09 & 37.99 & 39.26 & 39.23 & 37.63 & 38.62 & 38.03 & \textbf{39.67}
\\ 
\multirow{3}{*}{ISO = 1600}   
& 8 & 37.76 & 38.42 & 38.65 & 40.36 & 41.43 & 41.90 & 38.79 & 41.45 & 40.40 & \textbf{42.66}
\\ 
& 9 & 37.51 & 35.79 & 35.85 & 38.30 & 39.55 & 39.06 & 37.07 & 38.76 & 36.86 & \textbf{40.04}
\\
\hline
\multirow{3}{*}{Nikon D800} 
& 10 & 35.05 & 38.36 & 38.56 & 39.01 & 38.91 & 40.03 & 35.45 & 40.09 & 38.75 & \textbf{40.21}
\\ 
\multirow{3}{*}{ISO = 3200}   
& 11 & 34.07 & 35.53 & 35.76 & 36.75 & 37.41 & 36.89 & 35.43 & \textbf{37.57} & 36.52 & 37.30
\\ 
& 12 & 34.42 & 40.05 & 40.59 & 39.06 & 39.39 & 41.49 & 34.98 & 41.10 & 38.42 & \textbf{42.02}
\\ 
\hline
\multirow{3}{*}{Nikon D800} 
& 13 & 31.13 & 34.08 & 34.25 & 34.61 & 34.80 & 35.47 & 31.12 & 34.11 & 34.13 & \textbf{36.19}
\\ 
\multirow{3}{*}{ISO = 6400}   
& 14 & 31.22 & 32.13 & 32.38  & 33.21 & 33.95 & 34.05 & 31.93 & 33.64 & 33.45 & \textbf{34.70}
\\ 
& 15 & 30.97 & 31.52 & 31.76 & 33.22 & 33.94 & 33.88 & 31.79 & 33.68 & 33.45 & \textbf{34.83}
\\
\hline
\rowcolor[rgb]{ .9,  .9,  .9}  
\textbf{Average} 
& - & 35.19 & 35.33 & 35.65 & 36.88 & 37.71 & 37.81 & 35.40 & 37.63 & 36.44 & \textbf{38.49}
\\
\hline
\end{tabular}
\end{center}
\vspace{-1mm}
\label{t3}
\vspace{-0mm}
\end{table*}

\begin{table*}[htp]
\vspace{-0mm}
\caption{SSIM results of different denoising methods on the $15$ cropped real-world noisy images used in \textbf{CC} dataset~\cite{crosschannel2016}.}
\begin{center}
\renewcommand\arraystretch{1}
\footnotesize
\begin{tabular}{c||c|cccccccccc}
\hline
\rowcolor[rgb]{ .85,  .9,  .95}
Camera Settings 
&
\#
&
\textbf{CBM3D}
&
\textbf{NI}
&
\textbf{NC}
&
\textbf{CC}
&
\textbf{\footnotesize MCWNNM} 
&
\textbf{TWSC}
&
\textbf{DnCNN+}
&
\textbf{FFDNet+}
&
\textbf{CBDNet}
&
\textbf{NLH}
\\
\hline
\multirow{3}{*}{\small{Canon 5D M3}}  
&1& 0.9778 & 0.9600 & 0.9689 & 0.9678 & 0.9807 & 0.9805 & 0.9613 & 0.9723 & 0.9613 & \textbf{0.9847}
\\
\multirow{3}{*}{ISO = 3200}   
&2& 0.9552 & 0.9308 & 0.9427 & 0.9359 & 0.9591 & 0.9394 & 0.9415 & 0.9514 & 0.9430 & \textbf{0.9612}
\\ 
&3& 0.9660 & 0.9463 & 0.9476 & 0.9478 & \textbf{0.9676} & 0.9460 & 0.9553 & 0.9614 & 0.9562 & 0.9667
\\
\hline
\multirow{3}{*}{Nikon D600} 
&4& 0.9330 & 0.9413 & 0.9497 & 0.9484 & 0.9558 & 0.9581 & 0.9442 & 0.9506 & 0.9478 & \textbf{0.9606}
\\ 
\multirow{3}{*}{ISO = 3200}   
&5& 0.9168 & 0.9251 & 0.9398 & 0.9293 & 0.9534 & 0.9575 & 0.9187 & 0.9544 & 0.9406 & \textbf{0.9581}
\\ 
&6& 0.9313 & 0.9481 & 0.9588 & 0.9799 & 0.9684 & 0.9849 & 0.9278 & 0.9833 & 0.9751 & \textbf{0.9858}
\\
\hline
\multirow{3}{*}{Nikon D800} 
&7& 0.9339 & 0.9506 & 0.9533 & 0.9575 & 0.9638 & 0.9671 & 0.9460 & 0.9590 & 0.9591 & \textbf{0.9709}
\\
\multirow{3}{*}{ISO = 1600}   
&8& 0.9383 & 0.9615 & 0.9591 & 0.9767 & 0.9683 & 0.9804 & 0.9547 & 0.9800 & 0.9781 & \textbf{0.9833}
\\ 
&9& 0.9277 & 0.9229 & 0.9406 & 0.9427 & 0.9537 & 0.9496 & 0.9170 & 0.9419 & 0.9183 & \textbf{0.9598}
\\
\hline
\multirow{3}{*}{Nikon D800} 
&10& 0.8866 & 0.9101 & 0.9466 & 0.9637 & 0.9629 & \textbf{0.9770} & 0.8897 & 0.9755 & 0.9540 & 0.9750
\\ 
\multirow{3}{*}{ISO = 3200}   
&11& 0.8928 & 0.9194 & 0.9309 & 0.9477 & 0.9510 & 0.9498 & 0.9221 & \textbf{0.9569} & 0.9476 & 0.9525
\\ 
&12& 0.8430 & 0.9001 & 0.9070 & 0.9544 & 0.9578 & \textbf{0.9790} & 0.8563 & 0.9753 & 0.9492 & 0.9783
\\ 
\hline
\multirow{3}{*}{Nikon D800} 
&13& 0.7952 & 0.9074 & 0.9024 & 0.9206 & 0.9187 & 0.9369 & 0.7889 & 0.9140 & 0.9179 & \textbf{0.9436}
\\ 
\multirow{3}{*}{ISO = 6400}   
&14& 0.8613 & 0.8649 & 0.9141 & 0.9369 & 0.9379 & 0.9501 & 0.8844 & 0.9370 & 0.9290 & \textbf{0.9563}
\\ 
&15& 0.8363 & 0.8295 & 0.8847 & 0.9118 & 0.9225 & 0.9223 & 0.8637 & 0.9190 & 0.9121 & \textbf{0.9320}
\\
\hline
\rowcolor[rgb]{ .9,  .9,  .9}  
\textbf{Average} 
&-& 0.9063 & 0.9212 & 0.9364 & 0.9481 & 0.9548 & 0.9586 & 0.9115 & 0.9555 & 0.9460 & \textbf{0.9647}
\\
\hline
\end{tabular}
\end{center}
\vspace{-0mm}
\label{t-CC-SSIM}
\vspace{-0mm}
\end{table*}

\begin{figure*}
\centering
\begin{minipage}[t]{0.184\textwidth}
\includegraphics[width=1\textwidth]{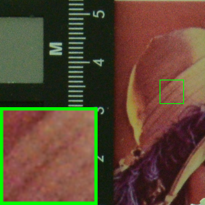}
\centering{\footnotesize (a) Noisy: 35.71/0.8839}
\includegraphics[width=1\textwidth]{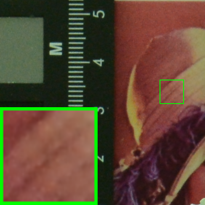}
\centering{\footnotesize (f) DnCNN+: 38.79/0.9547}
\end{minipage}
\begin{minipage}[t]{0.184\textwidth}
\includegraphics[width=1\textwidth]{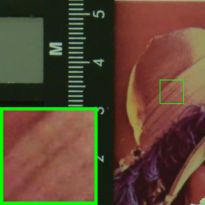}
\centering{\footnotesize (b) NC: 38.65/0.9591}
\includegraphics[width=1\textwidth]{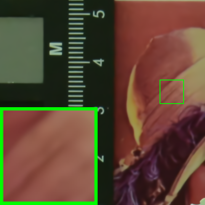}
\centering{\footnotesize (g) FFDNet+: 41.45/0.9800}
\end{minipage}
\begin{minipage}[t]{0.184\textwidth}
\includegraphics[width=1\textwidth]{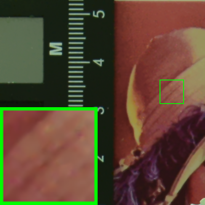}
\centering{\footnotesize (c) CC: 40.36/0.9767}
\includegraphics[width=1\textwidth]{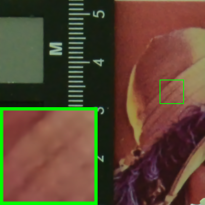}
\centering{\footnotesize (h) CBDNet: 40.40/0.9781}
\end{minipage}
\begin{minipage}[t]{0.184\textwidth}
\includegraphics[width=1\textwidth]{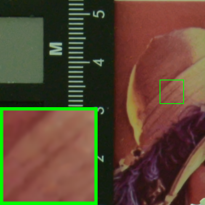}
\centering{\scriptsize (d) \scriptsize{MCWNNM: 41.43/0.9683}}
\includegraphics[width=1\textwidth]{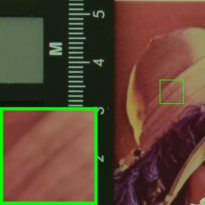}
\centering{\footnotesize (i) NLH: \textbf{42.66}/\textbf{0.9833}}
\end{minipage}
\begin{minipage}[t]{0.184\textwidth}
\includegraphics[width=1\textwidth]{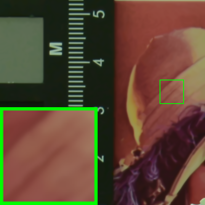}
\centering{\footnotesize (e) TWSC: 41.90/0.9804}
\includegraphics[width=1\textwidth]{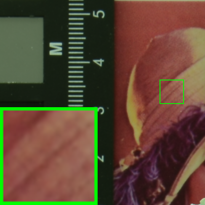}
\centering{\footnotesize (j) Mean Image}
\end{minipage}
\vspace{-0mm}
\caption{Comparison of denoised images and PSNR(dB)/SSIM by different methods on ``\textsl{Nikon D800 ISO=1600 2}" \cite{crosschannel2016}.}
\vspace{-0mm}
\label{f3}
\end{figure*}

\textbf{Comparison methods}.\ 
We compare the proposed NLH method with CBM3D \cite{cbm3d}, a commercial software Neat Image (NI) \cite{neatimage}, ``Noise Clinic'' (NC) \cite{noiseclinic}, Cross-Channel (CC) \cite{crosschannel2016}, MCWNNM \cite{mcwnnm}, TWSC \cite{twsc}.\
CBM3D can directly deal with color images, and the std of input noise is estimated by \cite{Chen2015ICCV}.\ 
For MCWNNM and TWSC, we use \cite{Chen2015ICCV} to estimate the noise std $\sigma_{c}$ ($c\in\{r,g,b\}$) for each channel and perform denoising accordingly.\ 
We also compare the proposed NLH method with DnCNN+ \cite{dncnn}, FFDNet+ \cite{ffdnet} and CBDNet \cite{cbdnet}, which are state-of-the-art convolutional neural network (CNN) based image denoising methods.\ 
FFDNet+ is a multi-scale extension of FFDNet \cite{ffdnet} with a manually selected uniform noise level map.\ 
DnCNN+ is based on the color version of DnCNN \cite{dncnn} for blind denoising, but fine-tuned with the results of FFDNet+ \cite{ffdnet}.\ 
Note that for FFDNet+ and DnCNN+, there is no need to estimate the noise std.\ 
For the three CNN based methods, we asked the authors to run the experiments for us.\ 
We also run the codes using our machine for speed comparisons.
\begin{figure*}[ht!]
\centering
\begin{minipage}[t]{0.184\textwidth}
\includegraphics[width=1\textwidth]{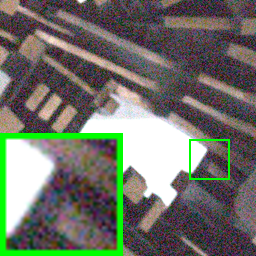}
\centering{\footnotesize (a) Noisy: 18.77/0.3015}
\includegraphics[width=1\textwidth]{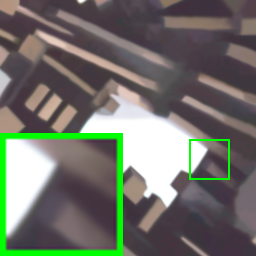}
\centering{\footnotesize (f) TWSC: \textbf{32.97}/0.9163}
\end{minipage}
\begin{minipage}[t]{0.184\textwidth}
\includegraphics[width=1\textwidth]{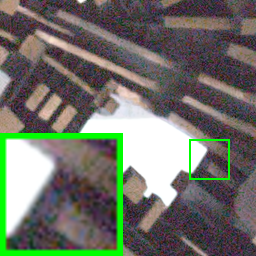}
\centering{\footnotesize (b) CBM3D: 23.95/0.5078}
\includegraphics[width=1\textwidth]{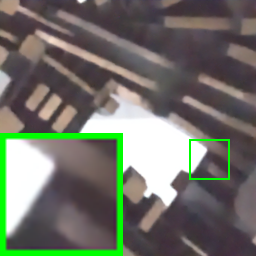}
\centering{\footnotesize (g) DnCNN+: 32.26/0.8906}
\end{minipage}
\begin{minipage}[t]{0.184\textwidth}
\includegraphics[width=1\textwidth]{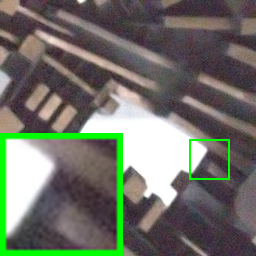}
\centering{\footnotesize (c) NI: 27.28/0.6330}
\includegraphics[width=1\textwidth]{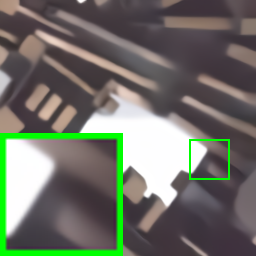}
\centering{\footnotesize (h) FFDNet+: 32.14/0.9162}
\end{minipage}
\begin{minipage}[t]{0.184\textwidth}
\includegraphics[width=1\textwidth]{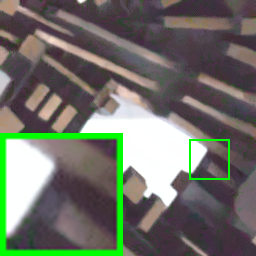}
\centering{\footnotesize (d) NC: 28.32/0.7186}
\includegraphics[width=1\textwidth]{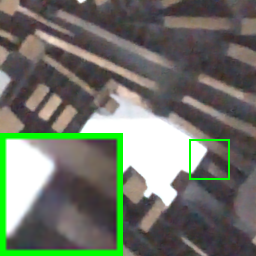}
\centering{\footnotesize (i) CBDNet: 31.40/0.8364}
\end{minipage}
\begin{minipage}[t]{0.184\textwidth}
\includegraphics[width=1\textwidth]{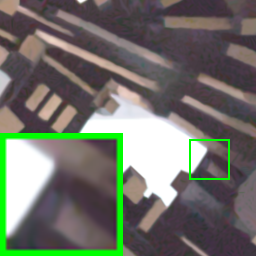}
\centering{\footnotesize (e) \scriptsize{MCWNNM: 31.74/0.8748}}
\includegraphics[width=1\textwidth]{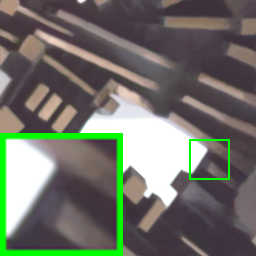}
\centering{\footnotesize (j) NLH: 32.85/\textbf{0.9202}}
\end{minipage}
\vspace{-1mm}
\caption{Comparison of denoised images and PSNR(dB)/SSIM by different methods on ``\textsl{0001\_18}", captured by a Nexus 6P \cite{dnd2017}.\ The ``ground-truth" image is not released, but PSNR(dB)/SSIM results are publicly provided on \href{https://noise.visinf.tu-darmstadt.de/benchmark/\#results_srgb}{DND's Website}.}
\vspace{-0mm}
\label{f4}
\end{figure*}

\begin{table*}[hbpt]
\vspace{-1mm}
\caption{Average results of PSNR(dB), SSIM, and CPU Time (in seconds) of different methods on 1000 cropped real-world noisy images in \textbf{DND} dataset \cite{dnd2017}. The GPU Time of DnCNN+, FFDNet+, and CBDNet are also reported in parentheses.}
\vspace{-2mm}
\begin{center}
\renewcommand\arraystretch{1}
\small
\begin{tabular}{c||cccccccccccc}
\Xhline{1pt}
\rowcolor[rgb]{ .85,  .9,  .95}
Metric
& \textbf{CBM3D}
& \textbf{NI}
& \textbf{NC}
& \textbf{MCWNNM}
& \textbf{TWSC}
& \textbf{DnCNN+}
& \textbf{FFDNet+}
& \textbf{CBDNet}
& \textbf{NLH}
\\
\hline
PSNR$\uparrow$
& 34.51 & 35.11 & 35.43 & 37.38 & 37.96 & 37.90 & 37.61 & 38.06 & \textbf{38.81}
\\
SSIM$\uparrow$ 
& 0.8507 & 0.8778 & 0.8841 & 0.9294 & 0.9416 & 0.9430 & 0.9415 & 0.9421& \textbf{0.9520}
\\
\hline
CPU (GPU) Time & 8.4 & \textbf{1.2} & 18.5 & 251.2 & 233.6 & 106.2 (0.05) & 49.9 (0.03) & 5.4 (0.40) & 5.3
\\
\hline
\end{tabular}
\end{center}
\vspace{-1mm}
\label{t4}
\vspace{-0mm}
\end{table*}

\textbf{Datasets and Results}.\ We evaluate our NLH on two benchmark datasets on real-world image denoising, i.e., the Cross-Channel (\textbf{CC}) dataset \cite{crosschannel2016} and the Darmstadt Noise Dataset (\textbf{DND}) \cite{dnd2017}. 

The \textbf{CC} dataset \cite{crosschannel2016} includes noisy images of 11 static scenes captured by Canon 5D Mark 3, Nikon D600, and Nikon D800 cameras.\ The real-world noisy images were collected under a controlled indoor environment.\ Each scene is shot 500 times using the same camera and settings.\ The average of the 500 shots is taken as the ``ground truth''.\ The authors cropped 15 images of size $512\times512$ to evaluate different denoising methods, as shown in Figure~\ref{f-cc}.\ The comparisons in terms of PSNR and SSIM are listed in Table~\ref{t3} and Table ~\ref{t-CC-SSIM}, respectively.\ It can be seen that, the proposed NLH method achieves the highest results on most images.\ Figure\ \ref{f3} shows the denoised images yielded by different methods on a scene captured by a Nikon D800 with ISO=1600.\ As can be seen, NLH also achieves better visual quality than other methods.\

The \textbf{DND} dataset \cite{dnd2017} includes 50 different scenes captured by Sony A7R, Olympus E-M10, Sony RX100 IV, and Huawei Nexus 6P.\ Each scene contains a pair of noisy and ``ground truth'' clean images.\ The noisy images are collected under higher ISO values with shorter exposure times, while the ``ground truth'' images are captured under lower ISO values with adjusted longer exposure times.\ For each scene, the authors cropped 20 bounding boxes of size $512\times512$, generating a total of 1000 test crops.\ The ``ground truth'' images are not released, but we can evaluate the performance by submitting the denoised images to the \href{https://noise.visinf.tu-darmstadt.de/benchmark/#results_srgb}{DND's Website}.\ In Table~\ref{t4}, we list the average PSNR (dB) and SSIM \cite{ssim} results of different methods.\ Figure~\ref{f4} shows the visual comparisons on the image ``\textsl{0001\_18}'' captured by a Nexus 6P camera.\ It can be seen that, the proposed NLH achieves higher PSNR and SSIM results, with better visual quality, than the other methods.\

\textbf{Speed}.\ We also compare the speed of all competing methods.\ All experiments are run under the Matlab 2016a environment on a machine with a quad-core 3.4GHz CPU and 8GB RAM.\ We also run DnCNN+, FFDNet+, and CBDNet on a Titan XP GPU.\ In Table~\ref{t4}, we also show the average run time (in seconds) of different methods, on the 1000 RGB images of size $512\times512$ in~\cite{dnd2017}.\ The fastest result is highlighted in bold.\ It can be seen that, Neat Image only needs an average of 1.2 seconds to process a $512\times512$ RGB image.\ The proposed NLH method needs $5.3$ seconds (using parallel computing), which is much faster than the other methods, including the patch-level NSS based methods such as MCWNNM and TWSC, the CNN based methods DnCNN+, FFDNet+, and CBDNet.\ The majority of time in the proposed NLH method is spent on searching similar patches, which takes an average of 2.8 seconds.\ Further searching similar pixels only takes an average of 0.3 seconds.\ This demonstrates that, the introduced pixel-level NSS prior adds only a small amount of calculation, when compared to its patch-level counterpart. 

\noindent

\begin{figure*}[ht!]
\centering
\subfigure[Extracting non-local similar pixels with synthetic Gaussian noise]{
\begin{minipage}[t]{0.6\linewidth}
\centering
\includegraphics[width=4in]{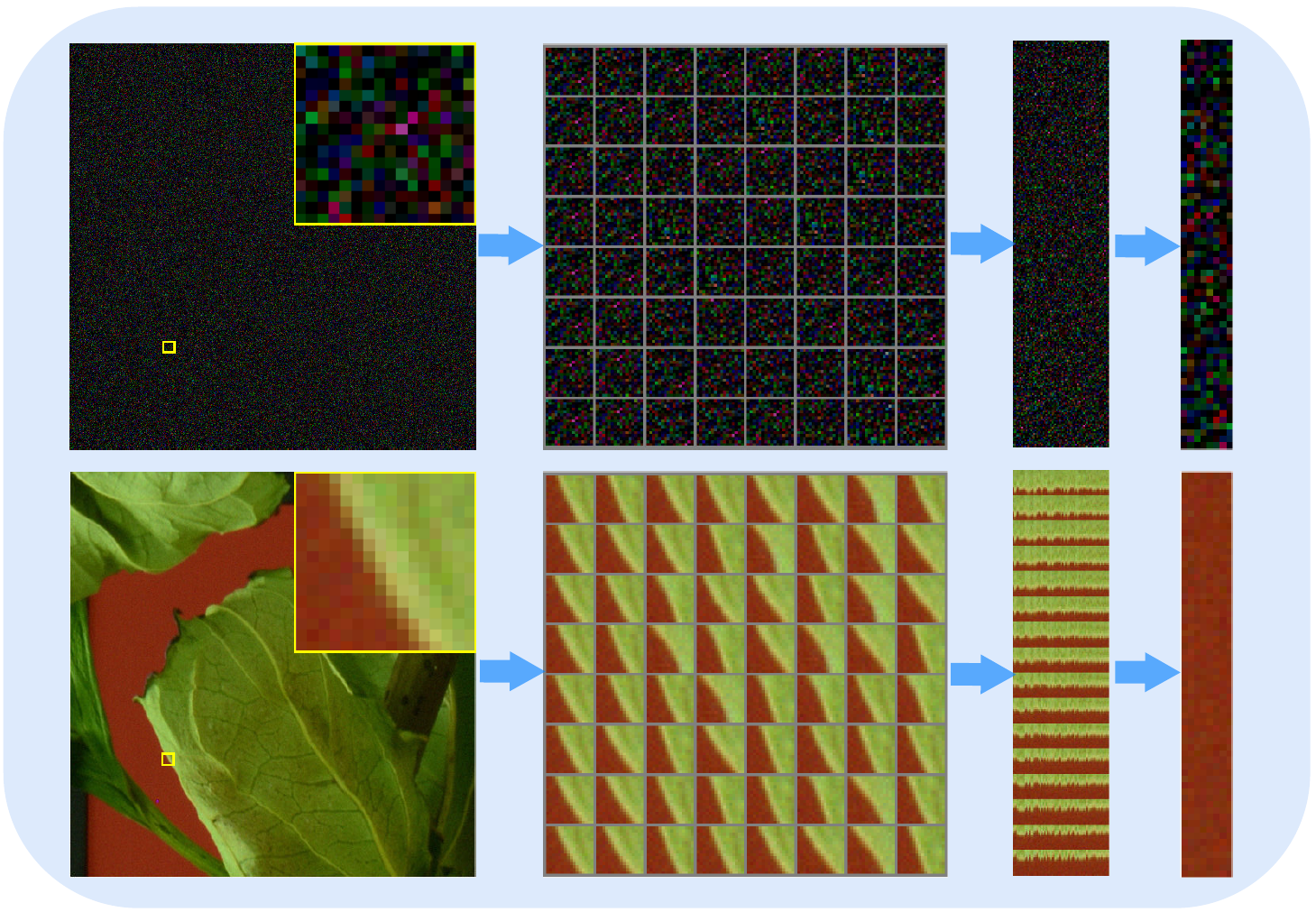}
\end{minipage}%
}%
\subfigure[Histogram of Gaussian noise]{
\begin{minipage}[t]{0.36\linewidth}
\centering
\includegraphics[width=2.5in]{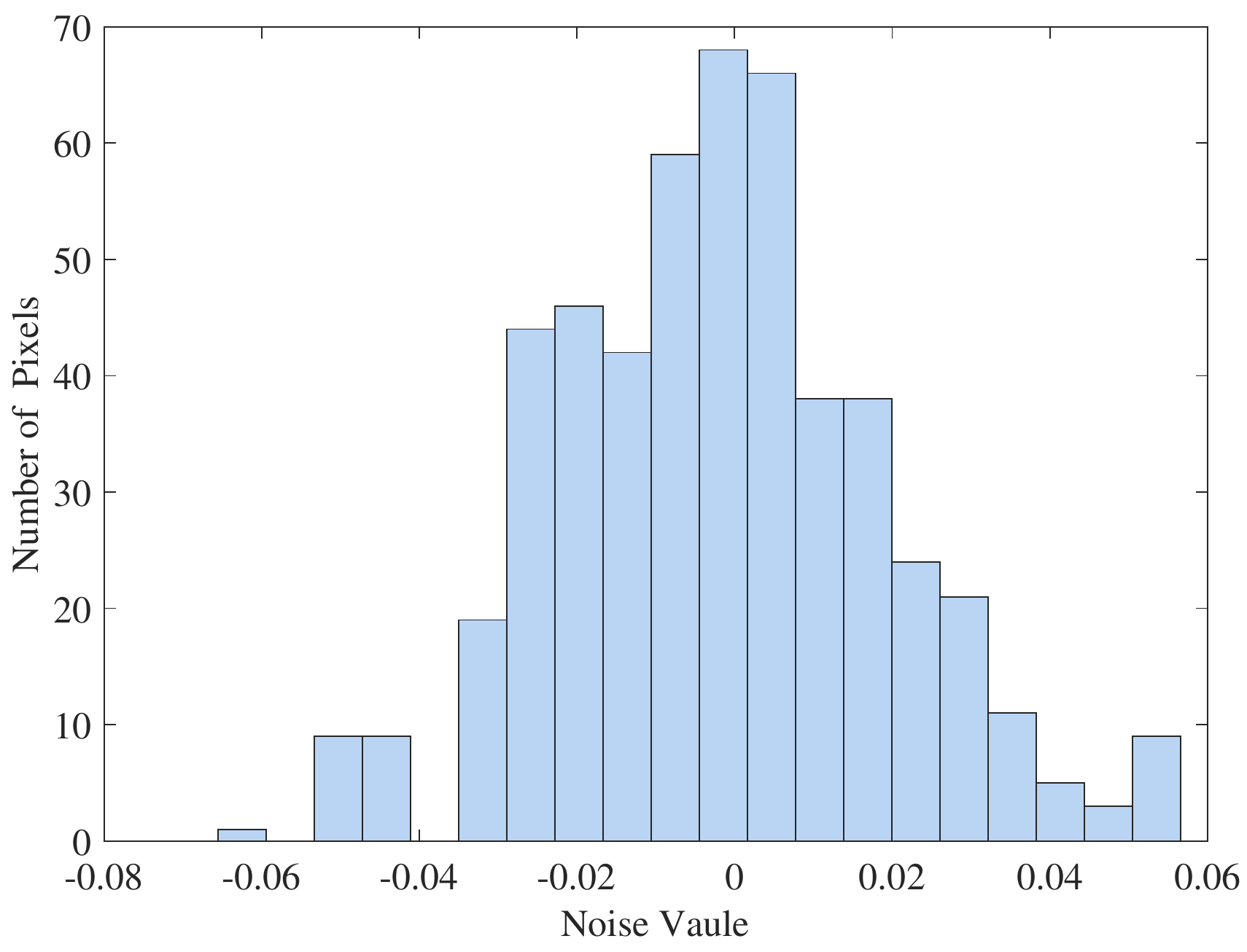}
\end{minipage}
}%
\\
\subfigure[Extracting non-local similar pixels with signal-dependent realistic noise]{
\begin{minipage}[t]{0.6\linewidth}
\centering
\includegraphics[width=4in]{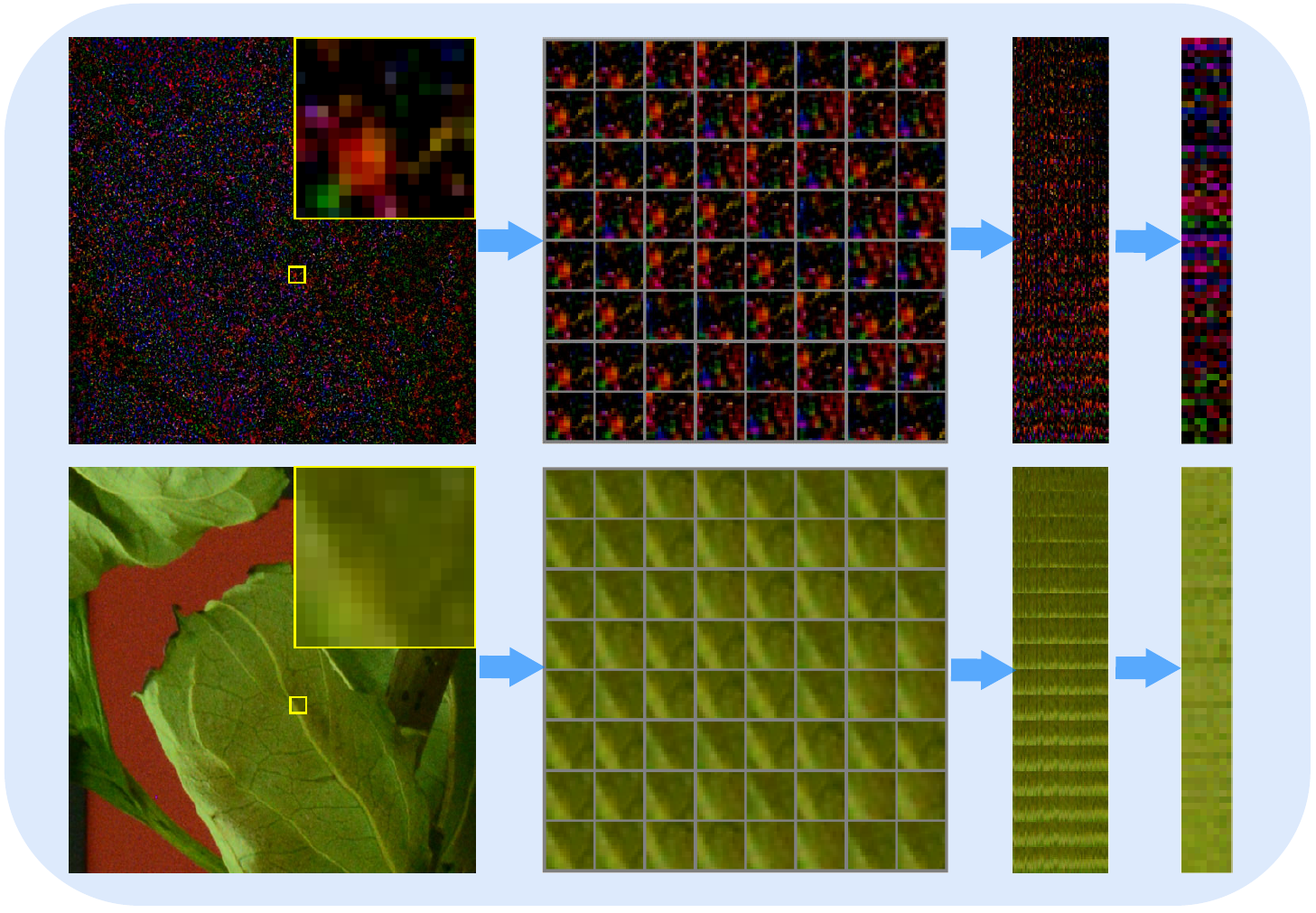}
\end{minipage}
}
\subfigure[Histogram of transformed realistic noise by NLH]{
\begin{minipage}[t]{0.36\linewidth}
\centering
\includegraphics[width=2.5in]{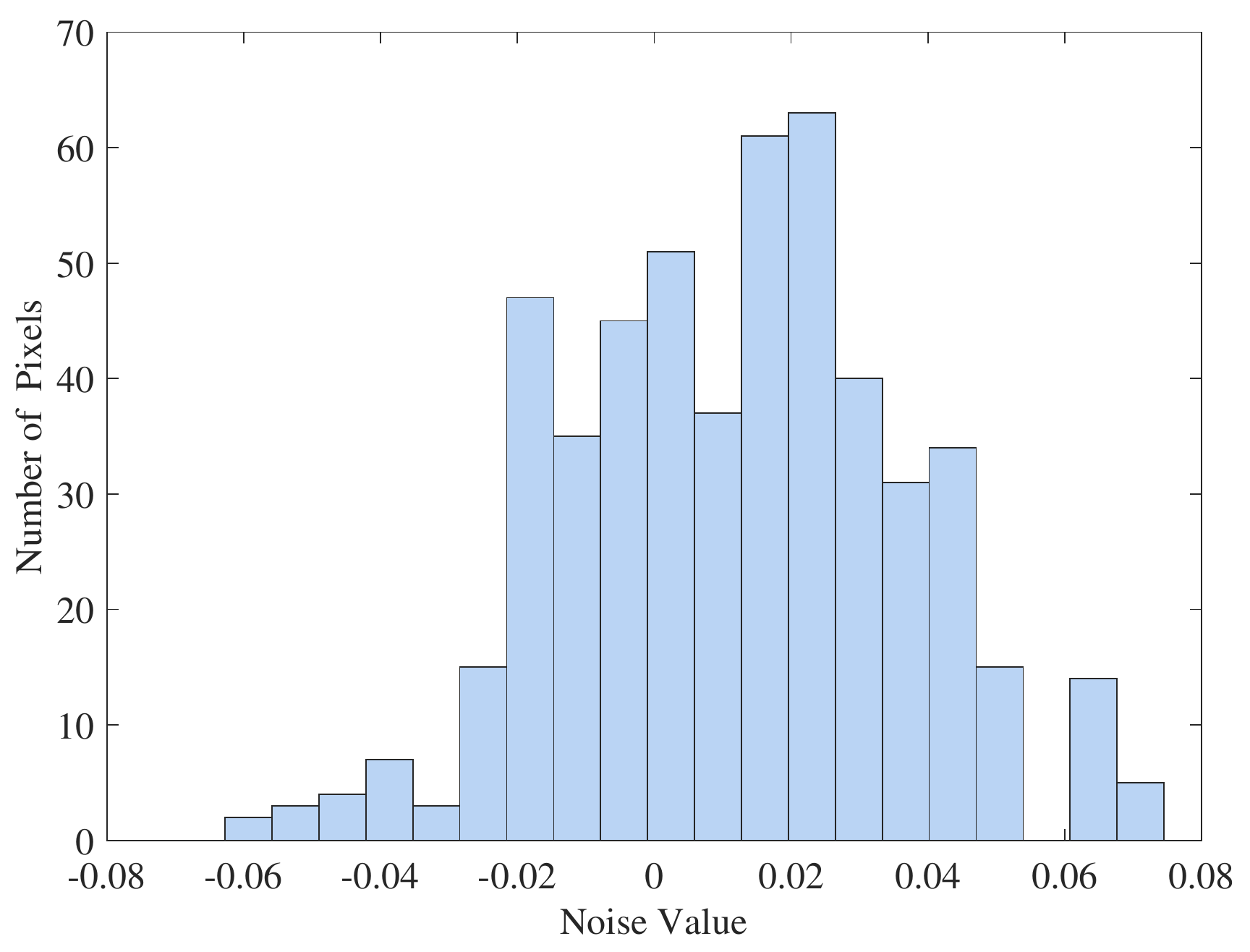}
\end{minipage}
}%
\vspace{-2mm}
\caption{Illustration of the extraction of non-local similar pixels in noisy images corrupted by synthetic Gaussian noise (a) and signal-dependent realistic noise (c).\
The histograms (b) and (d) of the noise in non-local similar pixels in (a) and (c), respectively, is close to Gaussian distributions, demonstrating that our NLH indeed transforms the signal-dependent realistic noise in image patches into quasi-Gaussian noise in non-local similar pixels.}
\label{fig:histogram}
\end{figure*}

\textbf{Discussion}.\
Our NLH achieves slightly improved results for gray-scale noisy image corrupted by AWGN noise, but is dramatically better on real-world noisy images when compared to the other methods~\cite{bm3d}, including the deep learning based methods~\cite{dncnn,cbdnet}.\ 
One the authors provide some kind of intuition on why the method 
our proposed non-local similar pixel searching scheme is feasible to transform the realistic noise, which is not Gaussian distributed~\cite{foi2008practical,dnd2017,twsc} to the quasi-Gaussian noise.\ 
To validate this point, we perform patch matching (block matching) followed by pixel matching (row matching) in a real-world noisy image from the CC dataset~\cite{crosschannel2016}.\
This process is illustrated in Figure~\ref{fig:histogram}. We add $\sigma$ = 5 Gaussian noise to the mean image (clean image) in CC dataset, then implement block matching and row matching to obtain a similar pixels group which can be seen in Figure~\ref{fig:histogram} (a), and we give the red channel noise histogram of the similar pixels group in Figure~\ref{fig:histogram} (b).\ 
On the other hand, we directly implement block matching and row patching on the real world image in Figure~\ref{fig:histogram} (c) and give its red channel noise histogram of similar pixels group in Figure~\ref{fig:histogram} (d).\ 
Because the signal-dependent noise color is mainly red, so comparing red channel histogram is much objective.\ 
We can see in two histograms that the noise value in two similar pixels groups are almost equal, this is the reason of adding $\sigma$ = 5 Gaussian noise.\ 
We observe that the realistic noise in Figure~\ref{fig:histogram} (c), which is usually not Gaussian distributed in images patches, is signal-dependent and can hardly be separated from the image patches.\
However, our proposed NLH is able to transform the signal dependent realistic noise to quasi-Gaussian noise.\ 
This ability of transforming realistic noise to quasi-Gaussian one makes our NLH very effective for realistic noise removal over previous image denoising methods, which are original designed for Gaussian noise removal or trained on the realistic noise different with the test ones.\ 
This is the key reason that why our NLH achieves dramatically better denoising performance on real-world noisy images, but similar performance on Gaussian noisy images, when compared to deep learning based approaches like DnCNN~\cite{dncnn}.

\vspace{-0mm}
\subsection{Validation of the Proposed NLH Method}
\label{sec:expvalid}
\vspace{-1mm}
We conduct more detailed examinations of our NLH to assess 1) the accuracy of pixel-level NSS \emph{vs.}\ patch-level NSS; 2) the contribution of the proposed pixel-level NSS prior for NLH on real-world image denoising; 3) the necessity of the two-stage framework; and 4) the individual influence of the 7 major parameters on NLH; 5) is the proposed noise estimation method or the proposed denoising algorithm contributes to the improvement of PSNR? 6) How the order of columns (or rows) influences our NLH on image denoising?

\noindent\textbf{1.\ Is pixel-level NSS more accurate than patch-level NSS?} 
To answer this question, we compute the average pixel-wise distances (APDs, the distance apportioned to each pixel) of non-local similar pixels and patches on the \textbf{CC} dataset~\cite{crosschannel2016}.\ From Table \ref{t5}, we can see that, on 15 mean images and 15 noisy images (normalized into $[0,1]$), the APDs of pixel-level NSS are smaller than those of patch-level NSS.\ In other words, pixel-level NSS is more accurate than the patch-level NSS on similarity measurements.
\begin{table}[htbp]
\vspace{-0mm}
\caption{Average pixel-wise distances of pixel-level NSS and patch-level NSS, on the 15 cropped mean images and corresponding noisy images in \textbf{CC} dataset~\cite{crosschannel2016}.}
\vspace{-3mm}
\begin{center}
\renewcommand\arraystretch{1}
\small
\begin{tabular}{c||ccc}
\Xhline{1pt}
\rowcolor[rgb]{ .85,  .9,  .95}
Aspect
&
\textbf{Mean Image}
&
\textbf{Noisy Image}
\\
\hline
\multicolumn{1}{r||}{Patch-level NSS} 
& $4.2\times10^{-4}$ & 0.0043
\\
\multicolumn{1}{r||}{Pixel-level NSS}
& $2.3\times10^{-4}$ & 0.0026
\\
\hline
\end{tabular}
\end{center}
\vspace{-2mm}
\label{t5}
\vspace{-0mm}
\end{table}

\noindent\textbf{2.\ Does pixel-level NSS prior contribute to image denoising?} Here, we study the contribution of the proposed pixel-level NSS prior.\ To this end, we remove the searching of pixel-level NSS in NLH.\ Thus we have a baseline: \textsl{w/o Pixel NSS}.\ From Table \ref{t6}, we observe a clear drop in PSNR (dB) and SSIM results over two datasets, which implies the effectiveness of the proposed pixel-level NSS prior.
\begin{table}[hbpt]
\vspace{-1mm}
\caption{Ablation study on the \textbf{CC}~\cite{crosschannel2016} and \textbf{DND}~\cite{dnd2017} datasets.\ We change one component at a time to assess its individual contributions to the proposed NLH method.}
\vspace{-3mm}
\begin{center}
\renewcommand\arraystretch{1}
\small
\begin{tabular}{c||c|c|c|c}
\Xhline{1pt}
\rowcolor[rgb]{ .85,  .9,  .95}
&
\multicolumn{2}{c|}{\textbf{CC~\cite{crosschannel2016}}}
&
\multicolumn{2}{c}{\textbf{DND~\cite{dnd2017}}}
\\
\rowcolor[rgb]{ .85,  .9,  .95}
\multirow{-2}{*}{Variant}
&
PSNR$\uparrow$
&
SSIM$\uparrow$
&
PSNR$\uparrow$
&
SSIM$\uparrow$
\\
\hline
\textbf{NLH} & 38.49 & 0.9647 & 38.81 & 0.9520
\\
\hline
\multicolumn{1}{l||}{\textsl{w/o Pixel NSS}} 
& 38.14 & 0.9602 & 38.27 & 0.9414
\\
\multicolumn{1}{l||}{\textsl{w/o \textbf{Stage 2}}}
& 37.64 & 0.9572 & 37.27 & 0.9355
\\
\hline
\end{tabular}
\end{center}
\vspace{-1mm}
\label{t6}
\vspace{-0mm}
\end{table}

\noindent\textbf{3.\ Is Stage 2 necessary?}\ We also study the effect of the \textbf{Stage 2} in NLH.\ To do so, we remove the \textbf{Stage 2} from NLH, and have a baseline:\ \textsl{w/o \textbf{Stage 2}}.\ From Table \ref{t6}, we can see a huge performance drop on two datasets.\ This shows that, the \textbf{Stage 2} complements the \textbf{Stage 1} with soft Wiener filtering, and is essential to the proposed NLH.\ 

\noindent\textbf{4.\ How each parameter influences NLH's denoising performance?}\ The proposed NLH mainly has 7 parameters (please see \S\ref{sec:expdetail} for details).\ We change one parameter at a time to assess its individual influence on NLH.\ Table~\ref{t7} lists the average PSNR results of NLH with different parameter values on \textbf{CC} dataset~\cite{crosschannel2016}.\ It can be seen that:\ 1) The variations of PSNR results are from 0.02dB (for iteration number $K$) to 0.16dB (for number of similar patches $m$), when changing individual parameters;\ 2) The performance on PSNR increases with increasing patch size $\sqrt{n}$, window size $W$, or iteration number $K$.\ For performance-speed tradeoff, we set $\sqrt{n}$$=$$7$, $W$$=$$40$, and $K$$=$$2$ in NLH for efficient image denoising;\ 3) The number of similar pixels $q$ is novel in NLH.\ To our surprise, even with $q$$=$$2$ similar pixels, NLH still performs very well, only drop 0.01dB on PSNR compared to case with $q$$=$$4$.\ However, with $q$$=$$8$$,$$16$, the performance of NLH decreases gradually.\ The reason is that, searching more (e.g., $16$) pixels in $7$$\times$$7$ patches may decrease the accuracy of pixel-level NSS, hence degrade the performance of NLH.\ Similar trends can be observed by changing the number of similar patches, i.e., the value of $m$.\ In summary, all the parametric analyses demonstrate that, NLH is very robust on real-world image denoising, as long as the 7 parameters are set in reasonable ranges.
\begin{table}[hbpt]
\vspace{-0mm}
\caption{PSNR (dB) of NLH with different parameters over the 15 noisy images in \textbf{CC} dataset~\cite{crosschannel2016}.\ We change one parameter at a time to assess its individual influence on NLH.}
\begin{center}
\renewcommand\arraystretch{1}
\small
\begin{tabular}{c||c|c|c|c|c||c}
\Xhline{1pt}
\multirow{2}{*}{$\sqrt{n}$}
& 
\cellcolor[rgb]{ .85,  .9,  .95}Value 
&
\cellcolor[rgb]{ .85,  .9,  .95}5 
&
\cellcolor[rgb]{ .85,  .9,  .95}6 
& 
\cellcolor[rgb]{ .85,  .9,  .95}7 
&
\cellcolor[rgb]{ .85,  .9,  .95}8 
& 
\cellcolor[rgb]{ .85,  .9,  .95}Margin
\\
& PSNR$\uparrow$
& 38.41 & 38.47 & 38.49 & 38.51 & 0.10
\\
\hline
\hline
\multirow{2}{*}{$W$}
& 
\cellcolor[rgb]{ .85,  .9,  .95}Value 
& 
\cellcolor[rgb]{ .85,  .9,  .95}20 
& 
\cellcolor[rgb]{ .85,  .9,  .95}30 
& 
\cellcolor[rgb]{ .85,  .9,  .95}40 
& 
\cellcolor[rgb]{ .85,  .9,  .95}50 
& 
\cellcolor[rgb]{ .85,  .9,  .95}Margin
\\
& PSNR$\uparrow$ 
& 38.39 & 38.43 & 38.49 & 38.51 & 0.12
\\
\hline
\hline
\multirow{2}{*}{$q$}
& 
\cellcolor[rgb]{ .85,  .9,  .95}Value
& 
\cellcolor[rgb]{ .85,  .9,  .95}2 
& 
\cellcolor[rgb]{ .85,  .9,  .95}4 
& 
\cellcolor[rgb]{ .85,  .9,  .95}8 
& 
\cellcolor[rgb]{ .85,  .9,  .95}16 
& 
\cellcolor[rgb]{ .85,  .9,  .95}Margin
\\
& PSNR$\uparrow$ 
& 38.48 & 38.49 & 38.47 & 38.43 & 0.06
\\
\hline
\hline
\multirow{2}{*}{$m$}
& 
\cellcolor[rgb]{ .85,  .9,  .95}Value 
& 
\cellcolor[rgb]{ .85,  .9,  .95}8 
& 
\cellcolor[rgb]{ .85,  .9,  .95}16 
& 
\cellcolor[rgb]{ .85,  .9,  .95}32 
& 
\cellcolor[rgb]{ .85,  .9,  .95}64 
& 
\cellcolor[rgb]{ .85,  .9,  .95}Margin 
\\
& PSNR$\uparrow$ 
& 38.33 & 38.49 & 38.48 & 38.43 & 0.16
\\
\hline
\hline
\multirow{2}{*}{$\tau$}
& \cellcolor[rgb]{ .85,  .9,  .95}Value 
& 
\cellcolor[rgb]{ .85,  .9,  .95}1.5 
& 
\cellcolor[rgb]{ .85,  .9,  .95}2 
& 
\cellcolor[rgb]{ .85,  .9,  .95}2.5 
& 
\cellcolor[rgb]{ .85,  .9,  .95}3 
& 
\cellcolor[rgb]{ .85,  .9,  .95}Margin
\\
& PSNR$\uparrow$ 
& 38.39 & 38.49 & 38.51 & 38.50 & 0.12
\\
\hline
\hline
\multirow{2}{*}{$K$}
& 
\cellcolor[rgb]{ .85,  .9,  .95}Value 
& 
\cellcolor[rgb]{ .85,  .9,  .95}2 
& 
\cellcolor[rgb]{ .85,  .9,  .95}3 
& 
\cellcolor[rgb]{ .85,  .9,  .95}4 
& 
\cellcolor[rgb]{ .85,  .9,  .95}5 
& 
\cellcolor[rgb]{ .85,  .9,  .95}Margin
\\
& PSNR$\uparrow$ 
& 38.49 & 38.51 & 38.51 & 38.51 & 0.02
\\
\hline
\hline
\multirow{2}{*}{$\lambda$}
& 
\cellcolor[rgb]{ .85,  .9,  .95}Value 
& 
\cellcolor[rgb]{ .85,  .9,  .95}0.2 
& 
\cellcolor[rgb]{ .85,  .9,  .95}0.4 
& 
\cellcolor[rgb]{ .85,  .9,  .95}0.6 
& 
\cellcolor[rgb]{ .85,  .9,  .95}0.8 
& 
\cellcolor[rgb]{ .85,  .9,  .95}Margin
\\
& PSNR$\uparrow$
& 38.46 & 38.47 & 38.49 & 38.49 & 0.03
\\
\hline
\end{tabular}
\end{center}
\vspace{-2mm}
\label{t7}
\vspace{-1mm}
\end{table}

\textbf{5. Is the proposed noise estimation method or the proposed denoising algorithm contributes to the improvement of the PSNR?} To anwser this question, we performed essential real-world image denoising experiments on the CC~\cite{crosschannel2016} dataset for the comparison methods such as CBM3D~\cite{cbm3d}, MCWNNM~\cite{mcwnnm}, and TWSC~\cite{twsc}, using our proposed noise estimation method (Eqn.\ (\ref{e4})).\ The PSNR and SSIM~\cite{ssim} results of these methods on the CC dataset are provided in Tables~\ref{t1-1}.\ We observe that, by using our proposed noise estimator, all these methods are improved with better denoising results on the CC dataset.\ However, even with the improvements, these methods still have a performance gap when compared with our NLH.\ This shows that accurate noise estimation is helpful, but not the most important component for state-of-the-art performance on image denoising.\ These results also demonstrate that the improvements of the denoising results by our NLH is not mainly from the proposed noise estimator, but is from the NLH denoising method itself.
\begin{table}[t]
\begin{center}
\caption{PSNR (dB) and SSIM results by different methods with the noise estimated by~\cite{Chen2015ICCV} and our noise estimation method (\ref{e4}) on the CC dataset~\cite{crosschannel2016}. %
}
\renewcommand\arraystretch{1}
\footnotesize
\begin{tabular}{c||ccccc}
\Xhline{1pt}
\rowcolor[rgb]{ .85,  .9,  .95}
Metric
&
Estimator
&
\textbf{CBM3D}
&
\textbf{MCWNNM}
&
\textbf{TWSC}
&
\textbf{NLH}
\\
\hline
\multirow{2}{*}{PSNR$\uparrow$ }
&
\cite{Chen2015ICCV}
& 35.19 & {37.71} & 37.81 & 38.33
\\
&
Our (\ref{e4})
& {35.65} & 37.68 & {37.82} & {38.49}
\\
\hline
\multirow{2}{*}{SSIM$\uparrow$ }
&
\cite{Chen2015ICCV}
& 0.9063 & 0.9548 & 0.9586 & 0.9622
\\
& 
Our (\ref{e4})
&{0.9211} & {0.9557} & {0.9588} & {0.9647}
\\
\hline
\end{tabular}
\end{center}
\vspace{-3mm}
\label{t1-1}
\end{table}

\begin{table}[htbp]
\caption{PSNR (dB) and SSIM results by different row order and column order in similar pixels matrices.}
\begin{center}
\renewcommand\arraystretch{1}
\small
\begin{tabular}{c|ccc}
\Xhline{1pt}
\rowcolor[rgb]{ .85,  .9,  .95}
Metric
&NLH&Switch Columns&Switch Rows
\\
\hline
PSNR$\uparrow$ &38.49 & 31.08 & 38.49 
\\
\hline
SSIM$\uparrow$ &0.9647 & 0.8861 & 0.9647 
\\
\hline
\end{tabular}
\end{center}
\vspace{-3mm}
\label{t-order}
\end{table}

6) How the order of columns (or rows) influences our NLH on image denoising?
To study this problem, we proposed two variants of the original NLH. 
The first variant is named ``Switch Columns'': for each similar patch matrix, we switch the reference patch (the first column) with its least similar column (the last column), while keeping the order of rows fixed.
The second variant is named ``Switch Rows'':
for each similar pixel matrix, we switch the reference row in the first row with the least similar row in the last row, while keeping the order of columns fixed.
As show in Table~\ref{t-order}, on the the CC dataset~\cite{crosschannel2016}, the variant of ``Switch Rows'' achieves close PSNR and SSIM~\cite{ssim} results with those of the original NLH.
However, the variant of ``Switch Column'' suffers from significant performance drop when compared to the original NLH. 
The key lies on the number of similar columns (or rows) in each similar patch (or pixel) matrix.
On one hand, the number of similar rows in our NLH is small (4 in stage one and 8 in stage two).
Thus in most cases the rows of pixels could be very similar to each other, and switching the rows does not influence little on the final results.
On the other, the number of similar patches is relatively large (16 in stage one and 64 in stage two) to exploit the non-local self similar property of natural images. 
Therefore, it is likely that some patches are not that similar to the reference one. 
Then, if we switch the reference patch in the first column with the least similar patch in the last column, the denoising results degrade significantly.

\section{Conclusion}
\label{conclusion}

How to utilize the non-local self similarity (NSS) prior for image denoising is an open problem.\ 
In this paper, we attempted to utilize the NSS prior to a greater extent by lifting the patch-level NSS prior to the pixel-level NSS prior.\ 
With the pixel-level NSS prior, we developed an accurate noise level estimation method, based on which we proposed a blind image denoising method.\ 
We estimated the local signal intensities via non-local Haar (NLH) transform based bi-hard thresholding, and performed denoising accordingly by Wiener filtering based soft thresholding.\ 
Experiments on benchmark datasets demonstrated that, the proposed NLH method significantly outperforms previous hand-crafted and non-deep methods, and is still competitive with existing state-of-the-art deep learning based methods on real-world image denoising task.\ 
We will simplify the pipeline of our NLH in the future work.

\section{Appendix: Detailed horizontal/vertical LHWT transforms and their inverse transforms}
\label{sec:appendix}
The Haar transform in our NLH is used differently from the traditional Haar transform on images.\ The reasons are: 1) in our NLH the matrices of similar pixels for Haar transform are not square ones, while traditional Haar transform needs square images; 2) these matrices of similar pixels are relatively small.\ Therefore, traditional orthogonal Haar transform used on image-level transform is not suitable for the small matrix-level scenarios in the proposed NLH.\ We employ an alternative lifting Haar transform, which is adaptive to our NLH.\

For each row $\bm{y}_{l}^{i}\in\mathbb{R}^{m}$ in the noisy patch matrix $\bm{Y}_l=[\bm{y}_{l,1},...,\bm{y}_{l,m}]\in\mathbb{R}^{n\times m}$, we select the $q$ ($q\ge2$) rows, i.e., $\{\bm{y}_l^{i_1},...,\bm{y}_l^{i_q}\}$ ($i_1=i$), in $\bm{Y}_l$ with the smallest Euclidean distances to $\bm{y}_l^{i}$, and stack the similar pixel rows as a matrix $\bm{Y}_{l}^{iq}=[{\bm{y}_{l}^{i_1}}^{\top},...,{\bm{y}_l^{i_q}}^{\top}]^{\top}\in\mathbb{R}^{q\times m}$.\ $\bm{Y}_l^{q}$ can also be written column by column as $\bm{Y}_l^{q}=[\bm{y}_{l,1}^{q},...,\bm{y}_{l,m}^{q}]\in\mathbb{R}^{n\times m}$, where $\bm{y}_{l,j}^{q}$ contains selected $q$ rows in $\bm{y}_{l,j}$ ($j=1,...,m$).\ For simplicity, we ignore the indices $i,l$ and have $\bm{Y}^{q}=[{\bm{y}^{1}}^{\top},...,{\bm{y}^{q}}^{\top}]^{\top}\in\mathbb{R}^{q\times m}$.\ $\bm{Y}_l^{q}$ is written as $\bm{Y}^{q}$, and $\bm{y}_{l,j}$ is written as $\bm{y}_{j}$ ($j=1,...,m$).\ Hence, $\bm{Y}^{q}$ can be written column by column as $\bm{Y}^{q}=[\bm{y}_{1}^{q},...,\bm{y}_{m}^{q}]\in\mathbb{R}^{q\times m}$, where $\bm{y}_{j}^{q}$ contains selected $q$ rows in $\bm{y}_{j}$ ($j=1,...,m$).

The proposed NLH contains horizontal and vertical LHWT transforms.\ For both stages, we set $q=4$, $m=16$ in all experiments.\ We first perform a horizontal LHWT transform (i.e., $\bm{C}^{4}=\bm{Y}^{4}\bm{H}_{r}$ as described in Eqn.\ (5) in the main paper):
\vspace{-2mm}
\begin{equation}
\label{e-1}
\begin{split}
&
\bm{c}_{t}^{4}
=
\frac{1}{\sqrt{16}}
(
\sum_{j=1}^{8}\bm{y}_{j}^{4}
+
(-1)^{t-1}
\sum_{j=9}^{16}\bm{y}_{j}^{4}
)
,
\ 
\text{when}
\ 
t=1,2
;
\\
&
\bm{c}_{t}^{4}
=
\frac{1}{\sqrt{8}}
(
\sum_{j=8(t-3)+1}^{8(t-3)+4}\bm{y}_{j}^{4}
-
\sum_{j=8(t-3)+5}^{8(t-2)}\bm{y}_{j}^{4}
)
,
\ 
\text{when}
\ 
t=3,4
;
\\
&
\bm{c}_{t}^{4}
=
\frac{1}{\sqrt{4}}
(
\sum_{j=4(t-5)+1}^{4(t-5)+2}\bm{y}_{j}^{4}
-
\sum_{j=4(t-5)+3}^{4(t-5)+4}\bm{y}_{j}^{4}
)
,
\ 
\text{when}
\ 
t=5,...,8
;
\\
&
\bm{c}_{t}^{4}
=
\frac{1}{\sqrt{2}}
(
\bm{y}_{2(t-9)+1}^{4}
-
\bm{y}_{2(t-9)+2}^{4}
)
,
\ 
\text{when}
\ 
t=9,...,16
.
\end{split}
\vspace{-2mm}
\end{equation}
We stack the coefficient vectors together and form $\bm{C}^{4}=[\bm{c}_{1}^{4},...,\bm{c}_{16}^{4}]\in\mathbb{R}^{4\times16}$.\ Assume that $\bm{c}^{i}\in\mathbb{R}^{16}$ is the $i$-th row of $\bm{C}^{4}$, i.e., $\bm{C}^{4}=[{\bm{c}^{1}}^{\top},...,{\bm{c}^{4}}^{\top}]^{\top}\in\mathbb{R}^{4\times16}$, we then perform vertical LHWT transform (i.e., $\bm{\hat{C}}^{4}=\bm{H}_{l}\bm{C}^{4}$ as described in Eqn.\ (5) in the main paper):
\vspace{-1mm}
\begin{equation}
\label{e-2}
\begin{split}
\bm{\hat{c}}^{1}
=
\frac{1}{\sqrt{4}}
\sum_{i=1}^{4}
\bm{c}^{i}
,
\ 
\bm{\hat{c}}^{2}
=
\frac{1}{\sqrt{4}}
(
\sum_{i=1}^{2}\bm{c}^{i}
-
\sum_{i=3}^{4}\bm{c}^{i}
)
,
\\
\bm{\hat{c}}^{3}
=
\frac{1}{\sqrt{2}}
(
\bm{c}^{1}
-
\bm{c}^{2}
)
,
\ 
\bm{\hat{c}}^{4}
=
\frac{1}{\sqrt{2}}
(
\bm{c}^{3}
-
\bm{c}^{4}
)
.
\end{split}
\vspace{-1mm}
\end{equation}
Then we perform a trivial hard thresholding operation:
\vspace{-2mm}
\begin{equation}
\label{e-3}
\bm{\hat{C}}^{4}
=
\bm{\hat{C}}^{4}
\odot
\mathbb{I}_{
\left\{
|\bm{\hat{C}}^{4}|\ge\tau\sigma_{g}^{2}
\right\}
}
,
\vspace{-2mm}
\end{equation}
where $\odot$ means element-wise production, $\mathbb{I}$ is the indicator function, and $\tau$ is the threshold parameter.\ We also perform a structurally hard thresholding and completely set to $0$ all the coefficients in the high frequency bands of $\bm{\hat{C}}^{4}$:
\vspace{-2mm}
\begin{equation}
\label{e-4}
\bm{\hat{C}}^{4}(i,j)
=
\bm{\hat{C}}^{4}(i,j)
\odot
\mathbb{I}_{
\left\{
\text{if}\ i=1,2\ \text{or}\ j=1
\right\}
}
,
\vspace{-2mm}
\end{equation}
where $\bm{\hat{C}}^{4}(i,j)$ is the $i,j$-th entry of the coefficient matrices $\bm{\hat{C}}^{4}$, respectively.

After the two hard thresholding steps, we perform inverse vertical and horizontal LHWT transforms.\ For simplicity, we still use the definitions in Eqn.\ (\ref{e-2}).\ We first perform an inverse vertical LHWT transform (i.e., $\bm{\widetilde{C}}^{4}=\bm{H}_{il}\bm{\hat{C}}^{4}$ as described in Eqn.~(\ref{e8})):
\vspace{-1mm}
\begin{equation}
\label{e-5}
\begin{split}
&
\bm{\widetilde{c}}^{1}
=
\frac{1}{\sqrt{4}}
(
\bm{\hat{c}}^{1}+\bm{\hat{c}}^{2}
)
+
\frac{1}{\sqrt{2}}
\bm{\hat{c}}^{3}
,
\\
&
\bm{\widetilde{c}}^{2}
=
\frac{1}{\sqrt{4}}
(
\bm{\hat{c}}^{1}+\bm{\hat{c}}^{2}
)
-
\frac{1}{\sqrt{2}}
\bm{\hat{c}}^{3}
,
\\
&
\bm{\widetilde{c}}^{3}
=
\frac{1}{\sqrt{4}}
(
\bm{\hat{c}}^{1}-\bm{\hat{c}}^{2}
)
+
\frac{1}{\sqrt{2}}
\bm{\hat{c}}^{4}
,
\\
&
\bm{\widetilde{c}}^{4}
=
\frac{1}{\sqrt{4}}
(
\bm{\hat{c}}^{1}-\bm{\hat{c}}^{2}
)
-
\frac{1}{\sqrt{2}}
\bm{\hat{c}}^{4}
.
\end{split}
\vspace{-6mm}
\end{equation}
We stack the rows of coefficients $\bm{\widetilde{c}}^{i}$ ($i=1,2,3,4$) together and form a matrix $\bm{\widetilde{C}}^{4}=[(\bm{\widetilde{c}}^{1})^{\top},...,(\bm{\widetilde{c}}^{4})^{\top}]^{\top}\in\mathbb{R}^{4\times16}$.\ Assume that $\bm{\widetilde{c}}_{j}^{4}\in\mathbb{R}^{4}$ is the $j$-th column of $\bm{\widetilde{C}}^{4}$, i.e., $\bm{\widetilde{C}}^{4}=[{\bm{\widetilde{c}}_{1}}^{4},...,{\bm{\widetilde{c}}_{16}^{4}}]\in\mathbb{R}^{4\times16}$, we then perform an inverse horizontal LHWT transform (i.e., $\bm{\widetilde{Y}}^{4}=\bm{\widetilde{C}}^{4}\bm{H}_{ir}$ as described in Eqn.~(\ref{e8})):
\begin{equation}
\label{e-6}
\tag{17}
\begin{split}
\bm{\widetilde{y}}_{1}^{4}
=
\frac{1}{\sqrt{16}}
(
\bm{\widetilde{c}}_{1}^{4}
+
\bm{\widetilde{c}}_{2}^{4}
)
+
\frac{1}{\sqrt{8}}
\bm{\widetilde{c}}_{3}^{4}
+
\frac{1}{\sqrt{4}}
\bm{\widetilde{c}}_{5}^{4}
+
\frac{1}{\sqrt{2}}
\bm{\widetilde{c}}_{9}^{4}
,
\\
\bm{\widetilde{y}}_{2}^{4}
=
\frac{1}{\sqrt{16}}
(
\bm{\widetilde{c}}_{1}^{4}
+
\bm{\widetilde{c}}_{2}^{4}
)
+
\frac{1}{\sqrt{8}}
\bm{\widetilde{c}}_{3}^{4}
+
\frac{1}{\sqrt{4}}
\bm{\widetilde{c}}_{5}^{4}
-
\frac{1}{\sqrt{2}}
\bm{\widetilde{c}}_{9}^{4}
,
\\
\bm{\widetilde{y}}_{3}^{4}
=
\frac{1}{\sqrt{16}}
(
\bm{\widetilde{c}}_{1}^{4}
+
\bm{\widetilde{c}}_{2}^{4}
)
+
\frac{1}{\sqrt{8}}
\bm{\widetilde{c}}_{3}^{4}
-
\frac{1}{\sqrt{4}}
\bm{\widetilde{c}}_{5}^{4}
+
\frac{1}{\sqrt{2}}
\bm{\widetilde{c}}_{10}^{4}
,
\\
\bm{\widetilde{y}}_{4}^{4}
=
\frac{1}{\sqrt{16}}
(
\bm{\widetilde{c}}_{1}^{4}
+
\bm{\widetilde{c}}_{2}^{4}
)
+
\frac{1}{\sqrt{8}}
\bm{\widetilde{c}}_{3}^{4}
-
\frac{1}{\sqrt{4}}
\bm{\widetilde{c}}_{5}^{4}
-
\frac{1}{\sqrt{2}}
\bm{\widetilde{c}}_{10}^{4}
,
\\
\bm{\widetilde{y}}_{5}^{4}
=
\frac{1}{\sqrt{16}}
(
\bm{\widetilde{c}}_{1}^{4}
+
\bm{\widetilde{c}}_{2}^{4}
)
-
\frac{1}{\sqrt{8}}
\bm{\widetilde{c}}_{3}^{4}
+
\frac{1}{\sqrt{4}}
\bm{\widetilde{c}}_{6}^{4}
+
\frac{1}{\sqrt{2}}
\bm{\widetilde{c}}_{11}^{4}
,
\\ 
\bm{\widetilde{y}}_{6}^{4}
=
\frac{1}{\sqrt{16}}
(
\bm{\widetilde{c}}_{1}^{4}
+
\bm{\widetilde{c}}_{2}^{4}
)
-
\frac{1}{\sqrt{8}}
\bm{\widetilde{c}}_{3}^{4}
+
\frac{1}{\sqrt{4}}
\bm{\widetilde{c}}_{6}^{4}
-
\frac{1}{\sqrt{2}}
\bm{\widetilde{c}}_{11}^{4}
,
\\
\bm{\widetilde{y}}_{7}^{4}
=
\frac{1}{\sqrt{16}}
(
\bm{\widetilde{c}}_{1}^{4}
+
\bm{\widetilde{c}}_{2}^{4}
)
-
\frac{1}{\sqrt{8}}
\bm{\widetilde{c}}_{3}^{4}
-
\frac{1}{\sqrt{4}}
\bm{\widetilde{c}}_{6}^{4}
+
\frac{1}{\sqrt{2}}
\bm{\widetilde{c}}_{12}^{4}
,
\\ 
\bm{\widetilde{y}}_{8}^{4}
=
\frac{1}{\sqrt{16}}
(
\bm{\widetilde{c}}_{1}^{4}
+
\bm{\widetilde{c}}_{2}^{4}
)
-
\frac{1}{\sqrt{8}}
\bm{\widetilde{c}}_{3}^{4}
-
\frac{1}{\sqrt{4}}
\bm{\widetilde{c}}_{6}^{4}
-
\frac{1}{\sqrt{2}}
\bm{\widetilde{c}}_{12}^{4}
,
\\
\end{split}
\end{equation}

\begin{equation}
\label{e-6}
\tag{17}
\begin{split}
\bm{\widetilde{y}}_{9}^{4}
=
\frac{1}{\sqrt{16}}
(
\bm{\widetilde{c}}_{1}^{4}
-
\bm{\widetilde{c}}_{2}^{4}
)
+
\frac{1}{\sqrt{8}}
\bm{\widetilde{c}}_{4}^{4}
+
\frac{1}{\sqrt{4}}
\bm{\widetilde{c}}_{7}^{4}
+
\frac{1}{\sqrt{2}}
\bm{\widetilde{c}}_{13}^{4}
,
\\ 
\bm{\widetilde{y}}_{10}^{4}
=
\frac{1}{\sqrt{16}}
(
\bm{\widetilde{c}}_{1}^{4}
-
\bm{\widetilde{c}}_{2}^{4}
)
+
\frac{1}{\sqrt{8}}
\bm{\widetilde{c}}_{4}^{4}
+
\frac{1}{\sqrt{4}}
\bm{\widetilde{c}}_{7}^{4}
-
\frac{1}{\sqrt{2}}
\bm{\widetilde{c}}_{13}^{4}
,
\\
\bm{\widetilde{y}}_{11}^{4}
=
\frac{1}{\sqrt{16}}
(
\bm{\widetilde{c}}_{1}^{4}
-
\bm{\widetilde{c}}_{2}^{4}
)
+
\frac{1}{\sqrt{8}}
\bm{\widetilde{c}}_{4}^{4}
-
\frac{1}{\sqrt{4}}
\bm{\widetilde{c}}_{7}^{4}
+
\frac{1}{\sqrt{2}}
\bm{\widetilde{c}}_{14}^{4}
,
\\ 
\bm{\widetilde{y}}_{12}^{4}
=
\frac{1}{\sqrt{16}}
(
\bm{\widetilde{c}}_{1}^{4}
-
\bm{\widetilde{c}}_{2}^{4}
)
+
\frac{1}{\sqrt{8}}
\bm{\widetilde{c}}_{4}^{4}
-
\frac{1}{\sqrt{4}}
\bm{\widetilde{c}}_{7}^{4}
-
\frac{1}{\sqrt{2}}
\bm{\widetilde{c}}_{14}^{4}
,
\\
\bm{\widetilde{y}}_{13}^{4}
=
\frac{1}{\sqrt{16}}
(
\bm{\widetilde{c}}_{1}^{4}
-
\bm{\widetilde{c}}_{2}^{4}
)
-
\frac{1}{\sqrt{8}}
\bm{\widetilde{c}}_{4}^{4}
+
\frac{1}{\sqrt{4}}
\bm{\widetilde{c}}_{8}^{4}
+
\frac{1}{\sqrt{2}}
\bm{\widetilde{c}}_{15}^{4}
,
\\ 
\bm{\widetilde{y}}_{14}^{4}
=
\frac{1}{\sqrt{16}}
(
\bm{\widetilde{c}}_{1}^{4}
-
\bm{\widetilde{c}}_{2}^{4}
)
-
\frac{1}{\sqrt{8}}
\bm{\widetilde{c}}_{4}^{4}
+
\frac{1}{\sqrt{4}}
\bm{\widetilde{c}}_{8}^{4}
-
\frac{1}{\sqrt{2}}
\bm{\widetilde{c}}_{15}^{4}
,
\\
\bm{\widetilde{y}}_{15}^{4}
=
\frac{1}{\sqrt{16}}
(
\bm{\widetilde{c}}_{1}^{4}
-
\bm{\widetilde{c}}_{2}^{4}
)
-
\frac{1}{\sqrt{8}}
\bm{\widetilde{c}}_{4}^{4}
-
\frac{1}{\sqrt{4}}
\bm{\widetilde{c}}_{8}^{4}
+
\frac{1}{\sqrt{2}}
\bm{\widetilde{c}}_{16}^{4}
,
\\ 
\bm{\widetilde{y}}_{16}^{4}
=
\frac{1}{\sqrt{16}}
(
\bm{\widetilde{c}}_{1}^{4}
-
\bm{\widetilde{c}}_{2}^{4}
)
-
\frac{1}{\sqrt{8}}
\bm{\widetilde{c}}_{4}^{4}
-
\frac{1}{\sqrt{4}}
\bm{\widetilde{c}}_{8}^{4}
-
\frac{1}{\sqrt{2}}
\bm{\widetilde{c}}_{16}^{4}
.
\end{split}
\vspace{-6mm}
\end{equation}
We stack $\{\bm{\widetilde{y}}_{j}^{4}\}_{j=1}^{16}$ together and form the denoised pixel matrix $\bm{\widetilde{Y}}^{4}=[\bm{\widetilde{y}}_{1}^{4},...,\bm{\widetilde{y}}_{16}^{4}]\in\mathbb{R}^{4\times16}$ in the first stage.\ We then aggregate all the denoised pixel matrices to form the denoised image.\ In the first stage, we perform the LHWT and inverse LHWT transforms for $K$ iterations.\ Note that we employ standard LHWT transforms without any modification.

{\small
\bibliographystyle{ieee}
\bibliography{NLH}
}
\vspace{-15mm}
\begin{IEEEbiography}[{\includegraphics[width=1in,height=1in,clip,keepaspectratio]{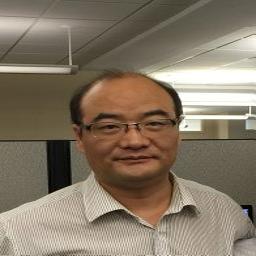}}]{Yingkun Hou} received his Ph. D degree from the School of Computer Science and Technology, Nanjing University of Science and Technology in 2012. He is currently an associate professor with the School of Information Science and Technology, Taishan University, Taian, China. His current research interests are in the areas of image processing, pattern recognition, and artificial intelligence.
\end{IEEEbiography}

\vspace{-15mm}
\begin{IEEEbiography}[{\includegraphics[width=1in,height=1in,clip,keepaspectratio]{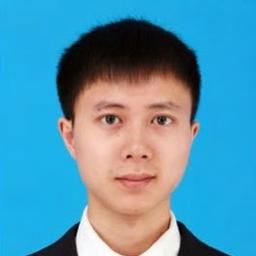}}]{Jun Xu} is an Assistant Professor in College of Computer Science, Nankai University, Tianjin, China.\
He received the B.Sc. and M. Sc. degrees in Mathematics in 2011 and 2014, respectively, from School of Mathematics Science, Nankai University, China.\ He received the Ph.D. degree in the Department of Computing, The Hong Kong Polytechnic University in 2018.\ He worked as a Research Scientist in Inception Institute of Artificial Intelligence (IIAI), Abu Dhabi, UAE.\ More information can be found on his homepage \url{https://csjunxu.github.io/}. 
\end{IEEEbiography}

\vspace{-15mm}
\begin{IEEEbiography}[{\includegraphics[width=1in,height=1in,clip,keepaspectratio]{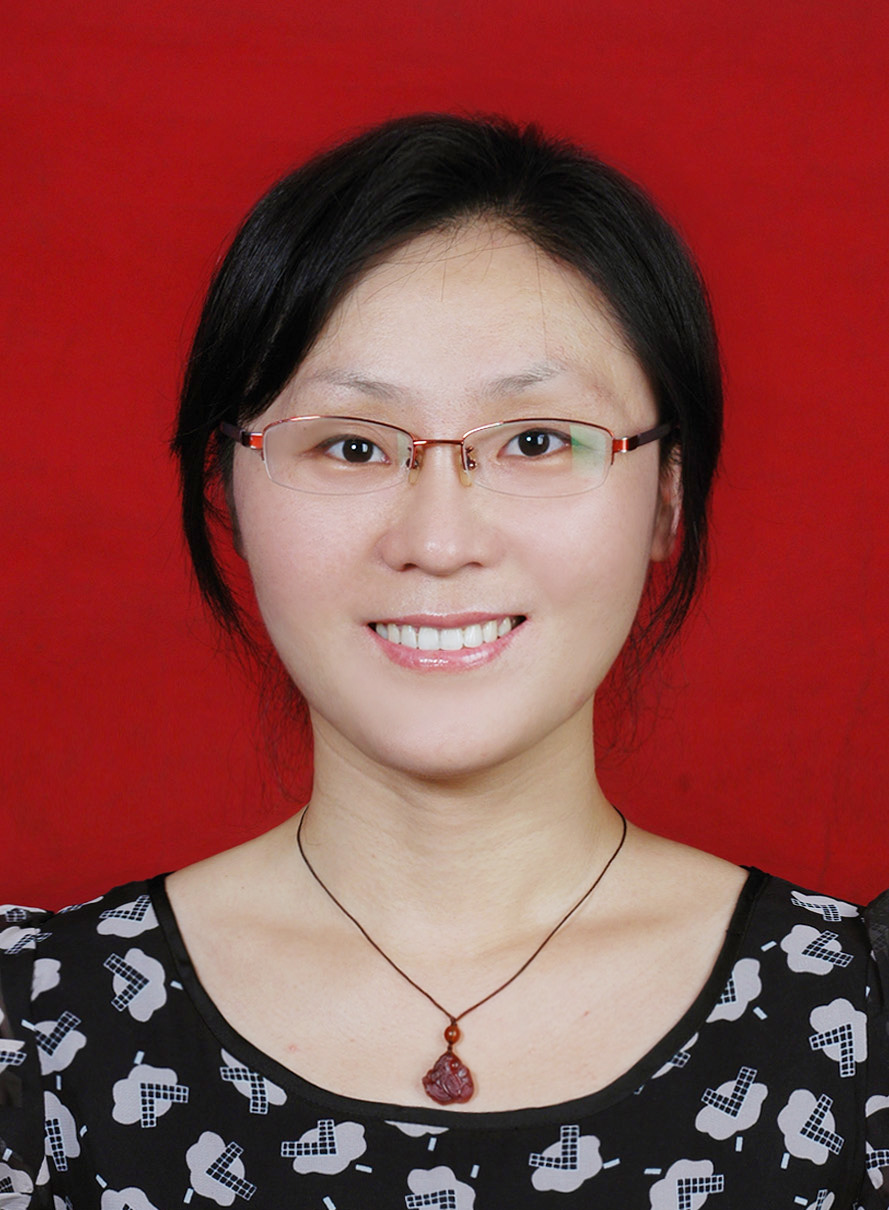}}]{Mingxia Liu} received the B.S. and M.S. degrees from Shandong Normal University, Shandong, China, in 2003 and 2006, respectively, and the Ph.D. degree from the Nanjing University of Aeronautics and Astronautics, Nanjing, China, in 2015. She is a Senior Member of IEEE (SM'19). Her current research interests include machine learning, pattern recognition, and medical image analysis.
\end{IEEEbiography}

\vspace{-15mm}
\begin{IEEEbiography}[{\includegraphics[width=1in,height=1in,clip,keepaspectratio]{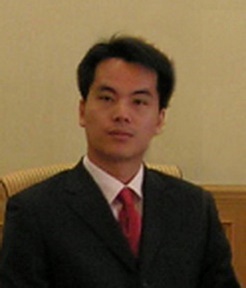}}]{Guanghai Liu} received his Ph. D degree from the School of Computer Science and Technology, Nanjing University of Science and Technology (NUST). He is currently a professor with the College of Computer Science and Information Technology, Guangxi Normal University in China. His current research interests are in the areas of image processing, pattern recognition, and artificial intelligence.
\end{IEEEbiography}

\vspace{-15mm}
\begin{IEEEbiography}[{\includegraphics[width=1in,height=1in,clip,keepaspectratio]{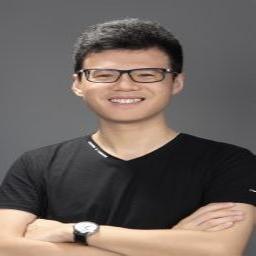}}]{Li Liu} is with Inception Institute of Artificial Intelligence (IIAI), Abu Dhabi, UAE.\
\end{IEEEbiography}

\vspace{-15mm}
\begin{IEEEbiography}[{\includegraphics[width=1in,height=1in,clip,keepaspectratio]{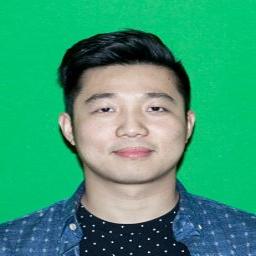}}]{Fan Zhu} is with Inception Institute of Artificial Intelligence (IIAI), Abu Dhabi, UAE.\
\end{IEEEbiography}

\vspace{-15mm}
\begin{IEEEbiography}[{\includegraphics[width=1in,height=1in,clip,keepaspectratio]{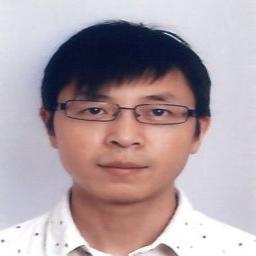}}]{Ling Shao} is the Executive Vice President and Provost of the Mohamed bin Zayed University of Artificial Intelligence. He is also the CEO and Chief Scientist of the Inception Institute of Artificial Intelligence (IIAI), Abu Dhabi, United Arab Emirates. His research interests include computer vision, machine learning, and medical imaging. He is a fellow of IAPR, IET, and BCS. 
\end{IEEEbiography}
\end{document}